\documentclass[journal]{IEEEtran}
\usepackage{cite}
\usepackage{threeparttable}
\usepackage{tabularx}
\usepackage{multicol}
\usepackage{multirow}
\usepackage{dcolumn}
\usepackage{booktabs}
\usepackage[pagebackref=true,breaklinks=true,letterpaper=true,colorlinks,bookmarks=false]{hyperref}

\newcommand{\PreserveBackslash}[1]{\let\temp=\\#1\let\\=\temp}
\newcolumntype{C}[1]{>{\PreserveBackslash\centering}p{#1}}
\newcolumntype{R}[1]{>{\PreserveBackslash\raggedleft}p{#1}}
\newcolumntype{L}[1]{>{\PreserveBackslash\raggedright}p{#1}}

\ifCLASSINFOpdf
   \usepackage[pdftex]{graphicx}
   \graphicspath{{pdf/}{jpeg/}}
   \DeclareGraphicsExtensions{.pdf,.jpeg,.png}
\else
   \usepackage[dvips]{graphicx}
   \graphicspath{{pdf/}}
   \DeclareGraphicsExtensions{.pdf}
\fi

\usepackage{caption}
\usepackage{subcaption}
\usepackage{amsmath}
\usepackage{array}

\begin{document}

\title{Twitter100k: A Real-world Dataset for Weakly Supervised Cross-Media Retrieval}

\author{Yuting~Hu, Liang Zheng, Yi Yang, and Yongfeng Huang
\thanks{Y.~Hu and Y.~Huang are with Tsinghua National Laboratory for Information Science and Technology, Department of Electronic Engineering, Tsinghua University, Beijing 100084, China. Email: huyt16@mails.tsinghua.edu.cn, yfhuang@tsinghua.edu.cn.}
\thanks{L.~Zheng and Y. Yang is with the Center of AI, University of Technology, Sydney. Email: liangzheng06@gmail.com, yi.yang@uts.edu.au.}
\thanks{The Twitter100k dataset together with the codes, feature files, training/testing set split and benchmarking results are available at http://ngn.ee.tsinghua.edu.cn/members/yuting-hu/.}}

\maketitle

\begin{abstract}
This paper contributes a new large-scale dataset for weakly supervised cross-media retrieval, named Twitter100k. Current datasets, such as Wikipedia, NUS Wide and Flickr30k, have two major limitations. First, these datasets are lacking in content diversity, \emph{i.e.}, only some pre-defined classes are covered. Second, texts in these datasets are written in well-organized language, leading to inconsistency with realistic applications. To overcome these drawbacks, the proposed Twitter100k dataset is characterized by two aspects: 1) it has 100,000 image-text pairs randomly crawled from Twitter and thus has no constraint in the image categories; 2) text in Twitter100k is written in informal language by the users.

Since strongly supervised methods leverage the class labels that may be missing in practice, this paper focuses on weakly supervised learning for cross-media retrieval, in which only text-image pairs are exploited during training. We extensively benchmark the performance of four subspace learning methods and three variants of the Correspondence AutoEncoder, along with various text features on Wikipedia, Flickr30k and Twitter100k. Novel insights are provided. As a minor contribution, inspired by the characteristic of Twitter100k, we propose an OCR-based cross-media retrieval method. In experiment, we show that the proposed OCR-based method improves the baseline performance.
\end{abstract}


\IEEEpeerreviewmaketitle

\section{Introduction}
\IEEEPARstart{C}{ross}-media retrieval has extensive applications. In this paper, we mainly discuss image-text retrieval. In this task, we aim to search relevant images (texts) from a large gallery that depict similar content with a text (image) query. The primary challenge in cross-media retrieval consists in eliminating the heterogeneity and mining the semantic correlation between modalities~\cite{kang2015learning,deng2016discriminative,yangyi2008,yangyi2008_2,yangyi2009,yangyi2012}. Another challenge is to speed up the process of finding relevant data to a query, which can be met by indexing techniques~\cite{song2013effective,wu2014sparse,zha2012interactive}. We focus on the first challenge in this work.

In the community, previous works can be categorized into two classes: weakly supervised and strongly supervised methods. For the former, only image-text pairs are available during training, \emph{e.g.}, multimodal DBM~\cite{pang2015deep}, DCCA~\cite{dcca2015}, MSAE~\cite{wang2014effective}, MSDS~\cite{MSDS2015} and CFA~\cite{CFA2003}; for the latter, class labels are provided for each modality, such as TINA~\cite{hua2016cross}, LCFS~\cite{LCFS2013}, JFSSL~\cite{JFSSL2016}, cluster-CCA~\cite{cluster-cca2014} and CAMH~\cite{liuruoyu2015}. This paper is concentrated on weakly supervised methods, which have critical research and application significance in realistic settings due to the expensive data annotation process~\cite{yangyi2016}.



This paper is motivated in two aspects. On the one hand, a majority of recent methods lay emphasis on leveraging fully supervised information such as the class labels. Usually, it is assumed that the training classes and testing classes are identical. However, this practice may be problematic in two aspects: 1) it seems a strong assumption that a query falls into a pre-defined class during training; 2) it might well be the case that labeled data is not available due to the annotation cost.

On the other hand, currently available cross-media datasets have some limitations.
First, they are usually deficient in content diversity. For example, the Pascal VOC 2012 dataset~\cite{pascal2012} has 20 different classes such as dog, horse, aeroplane \emph{etc}. However, retrieval involves multiple domains under realistic Internet circumstances. Retrieval methods trained on datasets of scanty domains may have difficulties in handling queries from unknown domains.
Second, texts in current text-image datasets are written in well-organized language, using standard grammar and spelling as well as proper words. However, texts may be written in a casual way and contain informal expressions in practice.
Third, each image in existing dataset is associated with tags or a highly-related text description. But in more realistic scenarios, images and texts are loosely correlated. Not all the words in texts have a visual interpretation in images.
Fourth, popular cross-media datasets consisting of sentences and images may be flawed in dataset scale, such as the IAPR TC-12 dataset~\cite{iaprtc2006} (20,000 samples) and the Wikipedia dataset~\cite{cca2010} (2,866 samples). The lack of data makes it challenging to evaluate the robustness of retrieval methods in large-scale galleries.

Considering the above-mentioned problems, this paper makes two major contributions. Our first contribution is the collection of a new large-scale cross-media dataset, named Twitter100k. It contains 100,000 image-text pairs collected from Twitter\footnote{twitter.com}. It is distinguished from existing datasets in two aspects: varied domains and informal text language. In result, this dataset provides a more realistic benchmark for cross-media analysis. Another contribution is that we provide extensive benchmarking experiments of the weakly supervised methods by testing the performance of four subspace learning methods~\cite{cca2010,blm1997,pls2006,gma2012} and three variants of Correspondence Autoencoder~\cite{corr-ae2014} on the new dataset along with the Wikipedia and Flickr30k datasets. Under the circumstances where class labels are unknown, pairs of texts and images are the only attainable training data. We propose to employ the cumulative match characteristic (CMC) curve rather than mean average precision (mAP) for accuracy evaluation.

As a minor contribution, inspired by the characteristics of Twitter100k that nearly 1/4 of the images contain texts which are highly correlated to the paired tweets, we make use of the texts in images and propose an OCR-based retrieval method. The effectiveness of the method is verified by the experiment.

The rest of this paper is organized as follows: we first present an overview on commonly used methods and datasets for cross-media retrieval in Section \uppercase\expandafter{\romannumeral2}. Then, we introduce the Twitter100k dataset in detail in Section \uppercase\expandafter{\romannumeral3}. Inspired by the characteristic of the new dataset, we proposed an OCR-based retrieval method in Section \uppercase\expandafter{\romannumeral4}. The benchmarking methods and evaluation protocol are described in Section \uppercase\expandafter{\romannumeral5}. The benchmarking results and experimenal analysis are presented in Section \uppercase\expandafter{\romannumeral6}. Finally, Section \uppercase\expandafter{\romannumeral7} concludes this paper.

\section{Related Work}
\subsection{Cross-media Retrieval Methods}
\textbf{Subspace learning methods.}
Subspace learning methods learn a common space for cross-media data, in which the similarity between the two modalities can be measured by L$_2$ distance, cosine distance, \emph{etc}. Canonical Correlation analysis (CCA)~\cite{cca2010} learns subspaces in which the correlation between the two modalities is maximized. Partial least square (PLS)~\cite{pls2006} learns latent vectors by maximizing the covariance between the two modalities. Bilinear Model (BLM)~\cite{blm1997} learns a set of basis in which data of the same content and different modality will project to the same coordinates. CCA, PLS and BLM are united in a framework called Generalized Multiview Analysis (GMA) by defining a joint optimization of two objective functions over two different vector spaces~\cite{gma2012}. A multi-view extension of Marginal Fisher Analysis (MFA) is derived, called generalized multiview marginal fisher analysis (GMMFA).

\textbf{Topic models.}
Topic models are widely applied to cross-media problem. Correspondence Latent Dirichlet Allocation (Corr-LDA) is proposed to effectively model the joint distribution of the two modalities and the conditional distribution of the text given the image~\cite{corrLDA2003}. A set of latent topics serves as latent variable and is shared between the two modalities. In topic-regression multi-modal Latent Dirichlet Allocation (tr-mmLDA) model~\cite{trmmLDA2010}, two separate sets of topics are learned and correlated by a regression module. Multi-modal Document Random Field (MDRF) model\cite{MDRF2011} defines a Markov random field over LDA topic model and learns topics shared across connected documents to encode the relations between different modalities.

\textbf{Deep learning methods.}
Some cross-media retrieval methods are based on deep learning. Deep Restricted Boltzmann Machine (Deep RBM) is utilized to model the joint representations for the two modalities by learning a probability density over the space of multimodal inputs~\cite{deep_rbm2012}. Deep Canonical Correlation Analysis (DCCA)~\cite{dcca2013} finds complex nonlinear transformations of two modalities such that the resulting representations are highly linearly correlated. A Relational Generative Deep Belief Nets (RGDBN) model~\cite{RG-DBN2013} computes the latent features for social media that best embed both the content and observed relationships by integrating the Indian buffet process into the modified Deep Belief Nets. Inspired by the efficiency of convolutional neural networks (CNN) for image~\cite{zhengliang2016} and text~\cite{txt-cnn2014}, a method called Deep and Bidirectional Representation Learning Model (DBRLM) uses two types of convolutional neural networks to represent text and image~\cite{he2016cross}. Correspondence Autoencoder (Corr-AE)~\cite{corr-ae2014} methods adopt two autoencoders to reconstruct the input text and image features while adding a connective layer to two hidden layers such that representation learning and common space learning can be accomplished in a single process.

\subsection{Cross-media Datasets}
We introduce five commonly used datasets in cross-media retrieval. A brief summary of these datasets is provided in Table \ref{dataset}.

\setlength{\tabcolsep}{5.4pt}
\begin{table}
\begin{center}
\caption{Summary of some popular cross-media datasets.}
\begin{tabular}{l|lllc}
\hline
 dataset & modality & \# pairs & avg. text len. & \# classes\\
\hline
 Wikipedia&img./text & 2,866 & 640.5 words &10\\
 Flickr30k& img./sentence & 31,783& 5 sentences & n.a\\
 IAPR TC-12&img./text& 20,000 & 23.1 words & n.a\\
 Pascal VOC'12&img./tags &11,540 &1.4 tags& 20\\
 NUS-WIDE&img./tags & 269,648&2.4 tags & 81\\
 \hline
\end{tabular}\label{dataset}
\end{center}
\end{table}

\begin{figure}[htbp]
  \includegraphics[width=1\linewidth]{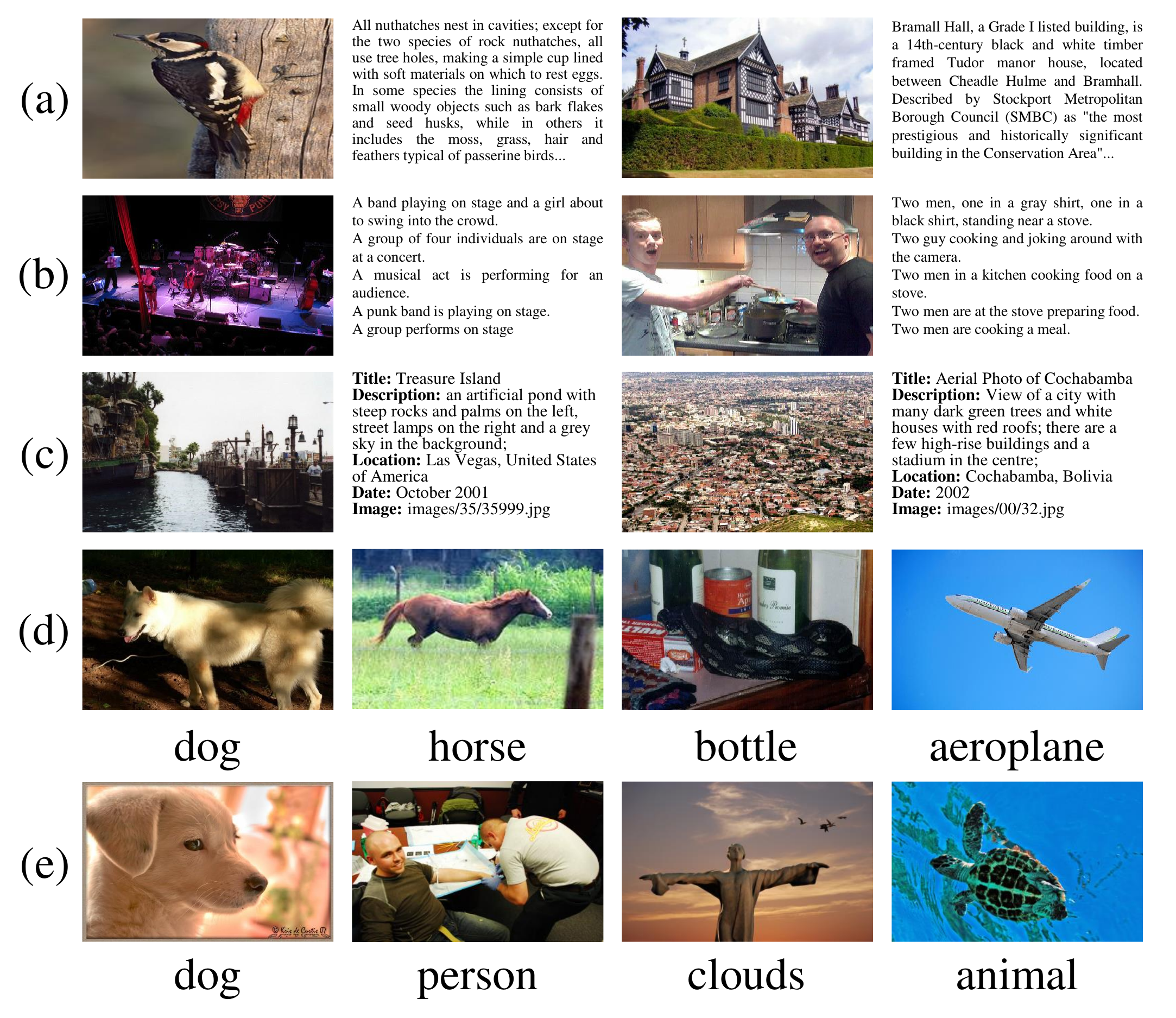}
  \caption{Examples in common-used cross-media datasets. (a)-(e) are examples in the Wikepedia, Flickr30k, IAPR TC-12, Pascal VOC 2012 and NUS-WIDE datasets, respectively.}
  \label{dataset_example}
\end{figure}

\textbf{The Wikipedia dataset.}
The Wikipedia dataset\footnote{http://www.svcl.ucsd.edu/projects/crossmodal/}~\cite{cca2010} contains 2,866 image-text pairs spread over 10 categories collected from Wikipedia's featured articles. The dataset is pruned to keep text that contains at least 70 words and the average text length is 640.5 words. The articles are determined by Wikipedia's editors and written in formal language. The shortcoming of this dataset is the insufficiency of the image-text pairs. We present two examples in this dataset in Fig. \ref{dataset_example}a.

\textbf{The Flickr30k dataset.}
The Flickr30k dataset\footnote{http://shannon.cs.illinois.edu/DenotationGraph/}~\cite{flickr2014} conprises 31,783 images from Flickr. Each image is associated with five descriptive sentences independently written by native English speakers from Mechanical Turk. Different annotators use different levels of specificity, from describing the overall situation to specific actions. The images and texts focus on people involved in everyday activities, events and scenes. No category information of this dataset is available. Examples in the dataset are given in Fig. \ref{dataset_example}b.

\textbf{The IAPR TC-12 dataset.}
The IAPR TC-12 dataset\footnote{http://imageclef.org/photodata}~\cite{iaprtc2006} is made up of 20,000 image-text pairs including domains of sports, actions, people, animals, cities, landscapes, \emph{etc}. The images are taken from locations around the world. Each text is composed of a title, a short description of the image, location and date of creation. The texts are written in English, German, Spanish or Portuguese. Each English text has 5.4 words in title and 23.1 words in the description averagely. We provide several examples in Fig. \ref{dataset_example}c.

\textbf{The Pascal VOC 2012 dataset.}
The Pascal VOC 2012\footnote{http://host.robots.ox.ac.uk/pascal/VOC/}~\cite{pascal2012} is a dataset designed for classification and detection tasks. It consists of 11,540 image-tag pairs in 20 different classes. Tags are the objects in the images. Each image has 1.4 tags on average and 7,567 images are labeled with only one tag. Examples in this dataset are show in Fig. \ref{dataset_example}d.

\textbf{The NUS-WIDE dataset.}
The NUS-WIDE dataset\footnote{http://lms.comp.nus.edu.sg/research/NUS-WIDE.htm}~\cite{NUSWIDE2009} contains 269,648 images and the associated tags from Flickr. Six types of low-level features are extracted from these images including color histogram, edge direction histogram, wavelet texture, block-wise color moments and bag of words based on SIFT descriptions. Ground-truth for 81 concepts can be used for evaluation. Each image holds 2.4 concepts on average. Some examples in this dataset are illustrated in Fig. \ref{dataset_example}e.

\section{Twitter100k: A Multi-Domain Dataset}
In this section, we introduce a multi-domain dataset called Twitter100k, which is comprised of 100,000 image-text pairs collected from Twitter.

\subsection{Dataset Collection}
Individual steps in data collection are described below.

\textbf{Seed user gathering.} To ensure the diversity of the collected data, we obtain seed users by sending queries to Twitter with various topic words, such as trip, meal, fitness, sports, \emph{etc}. These randomly selected users serve as seed to acquire more user candidates.

\textbf{User candidate generation.} A web spider is developed to crawl the accounts of the users who are following the seed users. This step iterates several times until we get a long list of user candidates. The domains covered by the users are further enlarged with the iteration.

\textbf{Tweet collection.} Another web spider collects tweets with the corresponding images by visiting the homepages of all the users in the candidate list. We find that around 1/3 of the tweets are companied with images.

\textbf{Data pruning.} The image-tweet pair is pruned under any of the following situations.
\begin{itemize}
\item
Messy codes in tweets;
\item
Tweets without words;
\item
Tweets not written in English;
\item
Reduplicate tweets with same ID;
\item
Error images.
\end{itemize}

\begin{figure*}[htbp]
  \centering
  \includegraphics[width=1\linewidth]{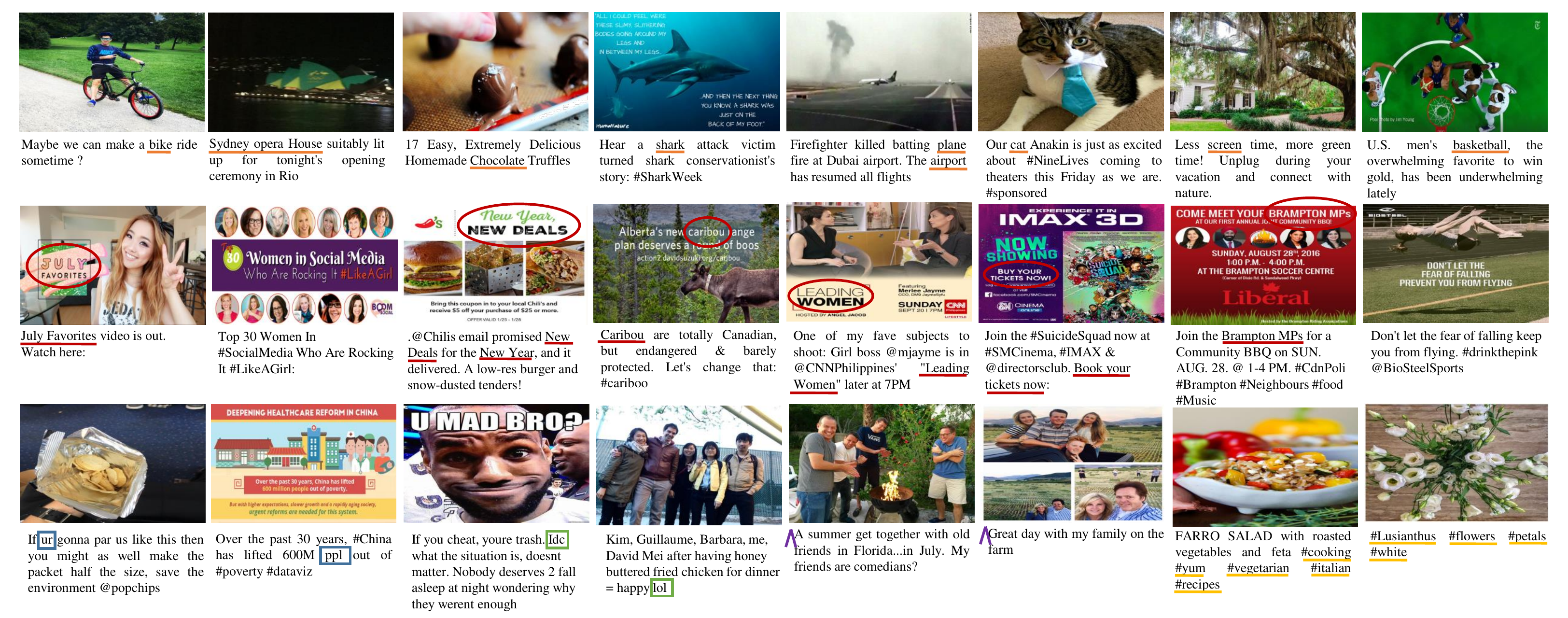}
  \caption{Examples in Twitter100k. In the first row are presented the examples whose images are visual interpretations of texts marked with orange lines. In the second row are presented the examples whose images contain words marked with red circles which are highly correlated to texts marked with red lines. In the third row are presented examples whose texts are written in informal language: abbreviations marked with blue panes, initialisms marked with green panes, omission of subject and verb marked with purple angles, hashtags marked with yellow lines.}
  \label{twitter100k_example} 
\end{figure*}

We finally obtain 100,000 image-text pairs in total. An image and text appearing in one piece of tweet are considered as a pair. Some examples in this new dataset are presented in Fig. \ref{twitter100k_example}.

\subsection{Dataset Characteristics}
The Twitter100k dataset is featured in the following five aspects. First, this dataset is collected from social media, hence it covers a wide range of domains, such as sport, architecture, food, animal, news, plant, person, poster and so forth.

Second, since informal language is usually used by Internet users when posting tweets, texts in Twitter100k are distinguished from other datasets in grammar and vocabularies.
\begin{itemize}
\item
\textbf{Abbreviations.} An abbreviation is a shortened form of a word or word group. Take the texts marked with blue panes in the third row of Fig. \ref{twitter100k_example} as example, \emph{ur} and \emph{ppl} are short for \emph{you are} and \emph{people}, respectively.
\item
\textbf{Initialisms.} An initialism is a term formed from the initial letter or letters of several words. In the texts marked with green panes in the third row of Fig. \ref{twitter100k_example}, \emph{idc} and \emph{lol} are initialisms for \emph{I don't care} and \emph{laughing out loud} in several.
\item
\textbf{Omission of subject and verb.} Both subject and verb are omitted when tweet is description or exclamation of an activity or something. The texts marked with purple angles in the third row of Fig. \ref{twitter100k_example} are examples.
\item
\textbf{Hashtags.} A hashtag is a type of label used on social network to indicate a specific theme. It consists of a hash character \# and a word or unspaced phrase. In the texts marked with yellow lines in the third row of Fig. \ref{twitter100k_example}, hashtags are highly correlated to the images.
\end{itemize}

Third, the correlation between the image and text is often very loose. In the examples in the first row of Fig. \ref{twitter100k_example}, some words in texts are direct descriptions of images while other words are out of semantic consistence and have no visual interpretations. Such is the case in practice. \par

Fourth, Twitter100k is a large-scale dataset, comprising 100,000 image-text pairs. A wealth of data can avoid overfitting during training. Moreover, it can be exploited to test the robustness of retrieval methods under massive data. \par

Last, approximately 1/4 of the images in this dataset contain text which are highly correlated to the paired tweets. As is presented by the examples in the second row of Fig. \ref{twitter100k_example}, some words in tweets are corresponding to the text located in images. In extreme cases, tweets can be identical to the text in images. To our knowledge, Twitter100k is the only cross-media dataset with this characteristic.

\subsection{Potential Application Scenarios}
The Twitter100k dataset provides a more realistic benchmark for cross-media retrieval. Since it is collected from Twitter, cross-media retrieval on this dataset is promising for application scenarios to be detailed below.
\begin{itemize}
\item
Social media platforms such as Twitter only provide pre-defined emoticons for users to choose when posting a tweet. In fact, by cross-media retrieval, it can be more convenient and interesting to extend the range of emoticons and recommend suitable images for users according to the contents of the tweets.
\item
Current content-based user recommendation and interest-group mining only take into consideration single-media data such as text. Adding correlation information between texts and images can improve the effectiveness of user recommendation systems and interest-group mining.
\end{itemize}

\section{Proposed OCR-based retrieval method for Twitter100k dataset}

It can be found out from the Twitter100k dataset that about 1/4 of the images contain text which may be correlated to the corresponding tweets and several of these images even involve no objects except text. It is challenging for current methods in cross-media retrieval to find the correlation between the tweets and these kind of images when optical character recognition (OCR) is not employed. Various features are used for image representation. For example, SIFT~\cite{sift2004} and color features are exploited in~\cite{zhengliang2014}; GIST~\cite{gist2001} and HOG~\cite{hog2005} are adopted in ~\cite{cca3v2014}. These features are effective in extracting color and shape of images, but defect in representing the words contained in the images. In order to tackle the problem, we propose an OCR-based cross-media retrieval method.

Our method consists of five components, which will be described below.

\textbf{OCR-text extraction.} We extract words on each image using python wrapper for Tesseract\footnote{Tesseract is an optical character recognition engine and considered as one of the most accurate open-source OCR engines. The code can be found at https://github.com/tesseract-ocr/tesseract}.

\textbf{OCR-text pruning.} As a revision of Tesseract, we prune the OCR-texts based on word frequency and text length. We first generate a vocabulary of the most frequent 5,000 words using the tweets in Twitter100k. After removing the words not in this vocabulary, we keep those OCR-texts which have at least two words remained. A total of 21,579 OCR-texts are obtained.

\textbf{Distance between tweet and OCR-text.} We adopt Jaccard Distance to measure the similarity between the tweet and OCR-text. The Jaccard Distance is defined as:
\begin{equation}
J(T,O)=\frac{M_{10}+M_{01}}{M_{10}+M_{01}+M_{11}},
\end{equation}
where $T$ and $O$ denote tweet and the OCR-text, respectively; M$_{11}$ is the number of words contained in both the tweet and OCR-text, M$_{10}$ is the number of words contained in tweet but not in OCR-text, and M$_{01}$ is the number of words contained in OCR-text but not in tweet.

\textbf{Distance in common subspace.} We use $T$ and $I$ to denote tweet and image. $C(T,I)$ represents the cosine distance between the two modalities in the common subspace learned by retrieval methods such as CCA, PLS, BLM, GMMFA and Corr-AE methods.

\textbf{Hybrid distance.} We define a hybrid distance between the tweet and image as follows:
\begin{equation}
dist(T,I)=\left\{\begin{matrix}
 {\alpha}J(T,O)+(1-\alpha)C(T,I),& Ind(I)=1\\
 C(T,I), & Ind(I)=0
\end{matrix}\right.
\end{equation}
where $\alpha$ is a weight parameter and $Ind(I)$ is a boolean-valued function to indicate whether image $I$ has a corresponding OCR-text. The influence of weight factor $\alpha$ is evaluated in the experiments.

\textbf{Ranking.} We rank the candidates in the gallery based on the hybrid distance between the query and candidate.

\section{Benchmarking Methods and Evaluation Protocol}
This paper mainly discuss the weakly supervised retrieval methods, in which only the co-occurrence infomation of the image and text is exploited. We will first describe some popular methods that will be adopted in the benchmarking experiment and then introduce the evaluation protocol.

\subsection{Compared Methods}
The cross-media retrieval methods we will compare in the paper are listed below.
\begin{itemize}
\item
\textbf{CCA.} Canonical correlation analysis (CCA)~\cite{cca2010} finds a basis of canonical components, \emph{i.e.}, directions, along which the data is maximally correlated.
\item
\textbf{BLM.} The bilinear model (BLM)~\cite{blm1997} can separate style and content by using singular value decomposition (SVD). It learns a shared subspace in which data of the same content and different modality is projected to the same coordinates.
\item
\textbf{PLS.} Partial least square (PLS)~\cite{pls2006} uses the least square method to correlate the subspaces of CCA in order to avoid information dissipation in the process of different modal correlations.
\item
\textbf{GMMFA.} Generalized multi-view marginal Fisher analysis (GMMFA)~\cite{gma2012} is a multi-view extension of Marginal Fisher Analysis. It tries to separate different-class and compress same-class samples in the feature space. We use GMMFA in a weakly supervised way by regarding every image-text pairs as an independent category.
\item
\textbf{Corr-AE.} Correspondence autoencoder (Corr-AE)~\cite{corr-ae2014} uses two autoencoders to reconstruct the input text and image features. It minimizes the weighted sum of the reconstruction loss and the distance between the hidden vectors of the two modalities.
\item
\textbf{cross Corr-AE.} Cross Corr-AE is a variant of Corr-AE. It takes one modality as input to reconstruct the other modality.
\item
\textbf{full Corr-AE.} Full Corr-AE is another variant of Corr-AE. It takes one modality as input to simultaneously reconstruct the two modalities.
\end{itemize}

Among the above-mentioned methods, CCA, BLM, PLS and GMMFA are subspace learning methods, while Corr-AE, cross Corr-AE and full Corr-AE are collectively called Corr-AE methods.

\subsection{Implement Details}
In this section, we describe the implementations and settings of the compared methods in detail.
\begin{itemize}
\item
\textbf{Subspace learning methods.} For subspace learning methods, we adopt the matlab implementation provided by~\cite{gma2012}\footnote{https://www.cs.umd.edu/~bhokaal/} to compute the linear projection matrix. Since no label information is considered in this paper, we set every text-image pair with an independent label. The options are set default.
\item
\textbf{Corr-AE methods.} For Corr-AE methods, we adopt the GPU-based python implementation provided by~\cite{corr-ae2014}\footnote{https://github.com/fangxiangfeng/deepnet} to compute the hidden vectors of the two modalities. According to the experiment results presented in~\cite{corr-ae2014}, 1024-dimensional hidden layer is employed. The weight factor of reconstruction error and correlation distance is set to 0.8, 0.2, 0.8 for Corr-AE, cross Corr-AE and full Corr-AE, respectively.
\end{itemize}

\subsection{Modality Representations}

\subsubsection{Text Representations}

\begin{itemize}
\item
\textbf{LDA feature.} For subspace learning methods, the representation of text is derived from a latent Dirichlet allocation (LDA)~\cite{lda2003} model. For each dataset, we first train a LDA model with 50 topics. Then each text is represented as a 50-dimensional LDA feature by the topic assignment probability distributions.
\item
\textbf{BoW feature.} For Corr-AE methods, text is represented by Bag-of-Word (BoW) feature. We first convert texts to lower case and remove stop-words. We adopt unigram model and select the most frequent 5,000 words to form a vocabulary. A 5000-dimensional BoW feature based on this vocabulary is generated for each text.
\item
\textbf{1024-dimensional WE-BoW feature.} Word embedding~\cite{wordembedding2013} (WE) is a language modeling and feature learning technique in natural language processing (NLP). It attempts to learn distributed representation of word, which is called word vector. Word vector contains semantic information of word and is applied to significantly improve many NLP applications~\cite{nlp2011}. Therefore, we propose to utilize a WE-BoW feature. A codebook of 1024 word vectors is first built with the k-means clustering algorithm using all the 400,000 300-dimensional word vectors pre-trained by GloVe\footnote{http://nlp.stanford.edu/projects/glove/}~\cite{glove2014}. Then word vectors in each text are quantized with this codebook, and text is represented by the L$_2$-normalized word vector histogram that results from this quantization.
\end{itemize}

\subsubsection{Image Representations}
\begin{itemize}
\item
\textbf{4096-dimensional CNN feature.} We first resize each image to 224*224 and extract the fc7 CNN feature using VGG16 model~\cite{vgg2014} with the implementation of CAFFE\footnote{http://caffe.berkeleyvision.org}~\cite{caffe2014}.
\end{itemize}

\subsection{Dataset Split}
In this paper, we benchmark the retrieval methods on the Wikipedia, Flickr30k and Twitter100k datasets. Each dataset is split into a training set, a validation set and a test set. The dataset split is described as follows:
\begin{itemize}
\item
\textbf{The Wikipedia dataset.} We use 2,173 image-text pairs for training and 500 pairs for testing. For Corr-AE methods, additional 193 pairs serve as validation set. All the data in test set is utilized as query.
\item
\textbf{The Flickr30k dataset.} The amounts of the training set and test set are both 15,000 image-text pairs. Extra 1,783 pairs are employed for validation in Corr-AE methods. We select 2,000 images and texts from the test set randomly to function as query. Since each image has 5 matched sentences, when taking image as query, the text gallery contains 75,000 sentences and any one of the 5 matched sentences is considered correct.
\item
\textbf{The Twitter100k dataset.} 50,000 and 40,000 image-text pairs are exploited for training and testing, respectively. For Corr-AE methods, 10,000 pairs are used as validation set. 2,000 images and texts are selected randomly from the test set to serve as query.
\end{itemize}

\subsection{Evaluation Metrics}
Since no pre-defined category labels are available, text and image in a pair are considered as a ground-truth match. That is, given a query text (image), only one ground-truth image (text) exists in the gallery. As a consequence, we take the following two evaluation metrics.
\begin{itemize}
\item
\textbf{CMC.} Cumulative match characteristic (CMC) is frequently used as a metric in the field of face recognition~\cite{cmc2000} and person re-identification~\cite{zhengliang2016mars}~\cite{zhengliang2016reid}. It measures how well an identification system ranks the identities in the enrolled database with respect to "unknown" probe image. For cross-media retrieval, CMC represents the expectation of finding the correct match in the top $n$ matches and can be described by a curve of average retrieval accuracy with respect to rank.
\item
\textbf{Mean rank.} Mean rank is the average of the ranks of the correct matches for a series of queries $Q$.
\begin{equation}
\textrm{Mean Rank}=\frac{1}{|Q|}\sum_{i=1}^{|Q|}{rank_i},
\end{equation}
where $rank_i$ refers to the rank position of the correct match for the $i$-th query.
\end{itemize}

\section{Experiment results}
In this section, we present the experiment results and discuss the impact of several factors on the retrieval performance, including datasets, the amount of training data, text features and retrieval methods.

\subsection{Comparison of Retrieval Methods on Various Datasets }
\begin{figure}[htbp]
  \begin{subfigure}{1.7in}
    \includegraphics[width=1.7in]{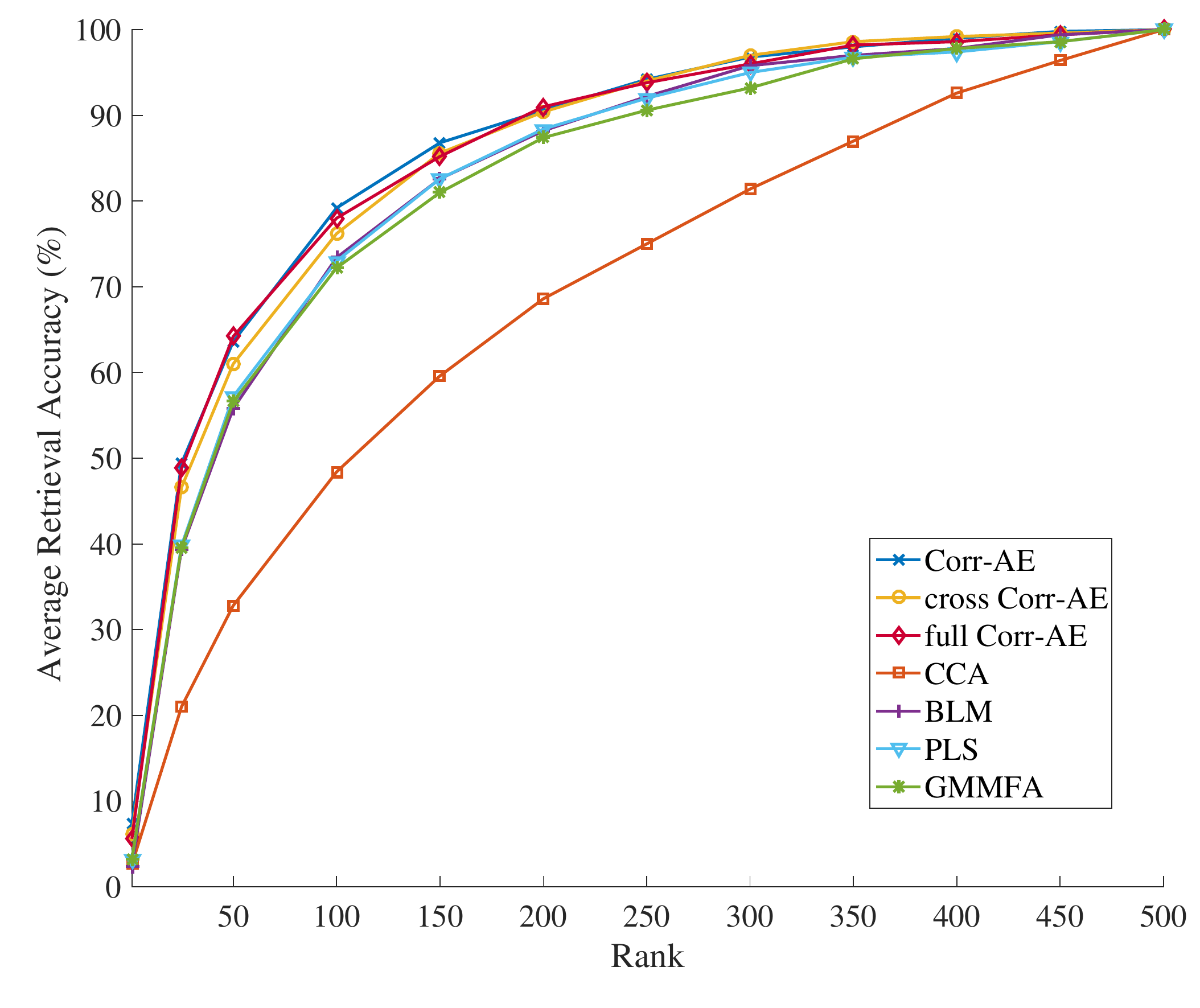}
    \vskip -2mm
    \caption*{Text to image retrieval}
  \end{subfigure}
  \begin{subfigure}{1.7in}
    \includegraphics[width=1.7in]{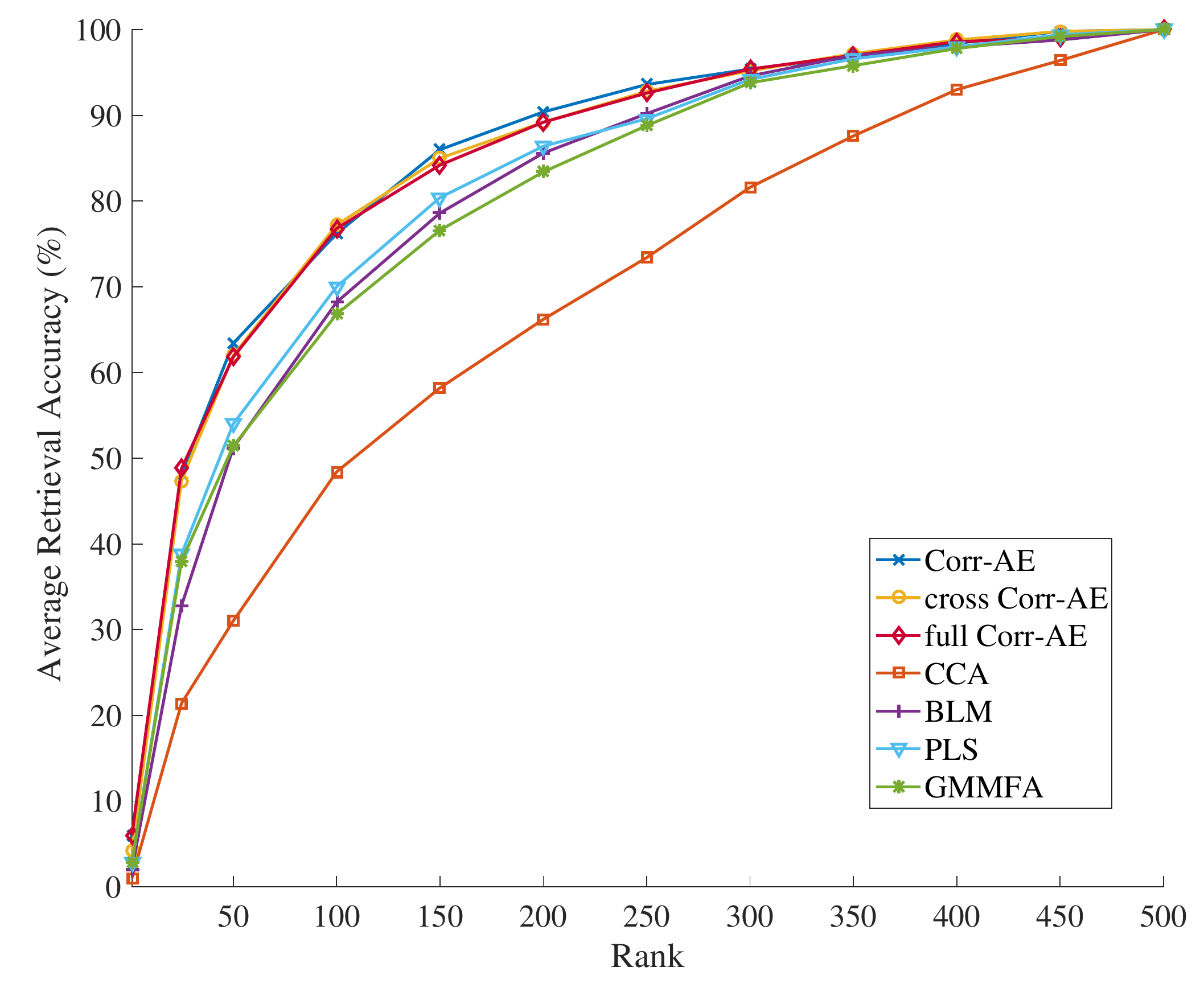}
    \vskip -2mm
    \caption*{Image to text retrieval}
  \end{subfigure}
  \begin{subfigure}{3.5in}
    \caption{Wikipedia}
  \end{subfigure}
  \hfill
  \begin{subfigure}{1.7in}
    \includegraphics[width=1.7in]{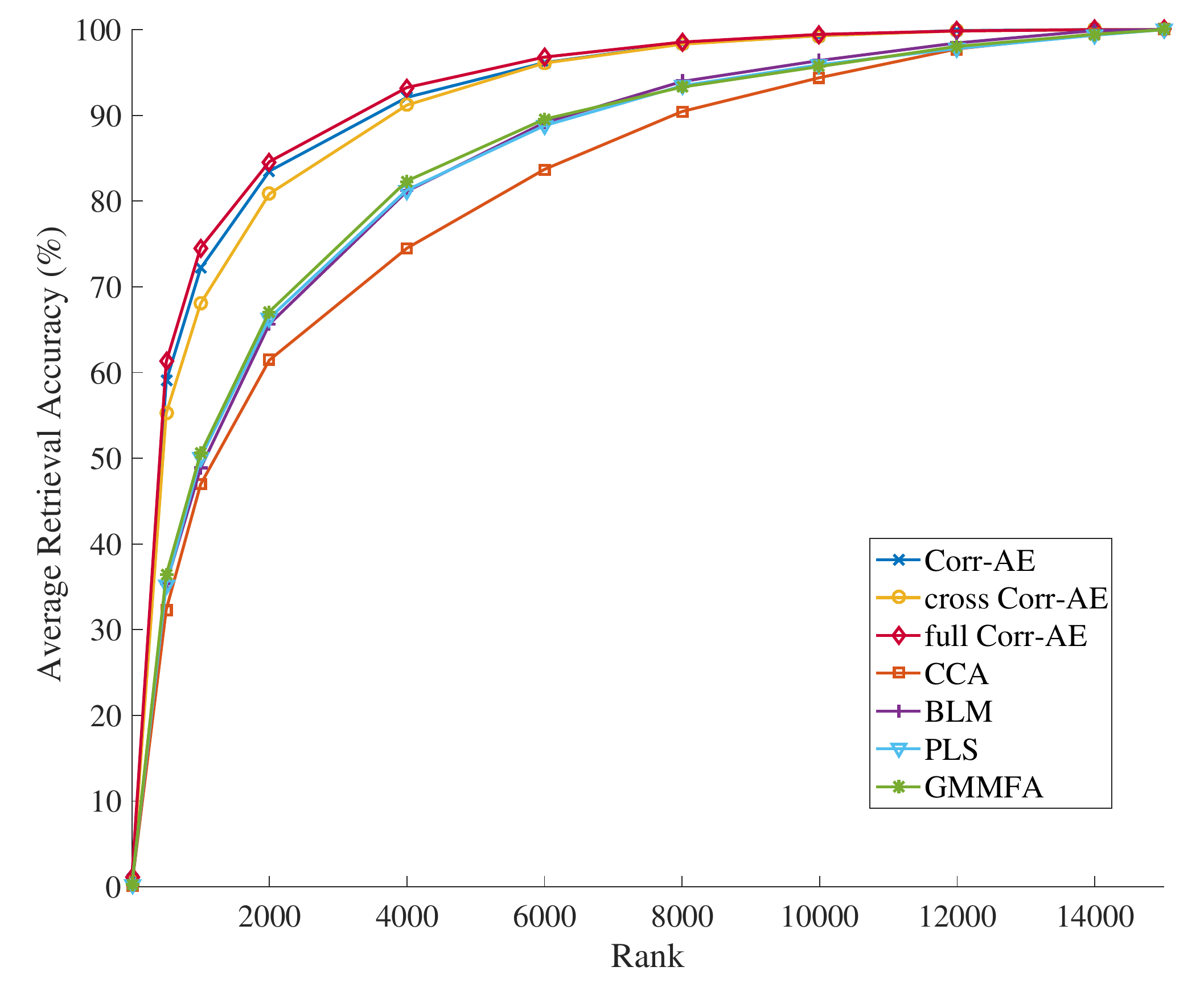}
    \vskip -2mm
    \caption*{Text to image retrieval}
  \end{subfigure}
  \begin{subfigure}{1.7in}
    \includegraphics[width=1.7in]{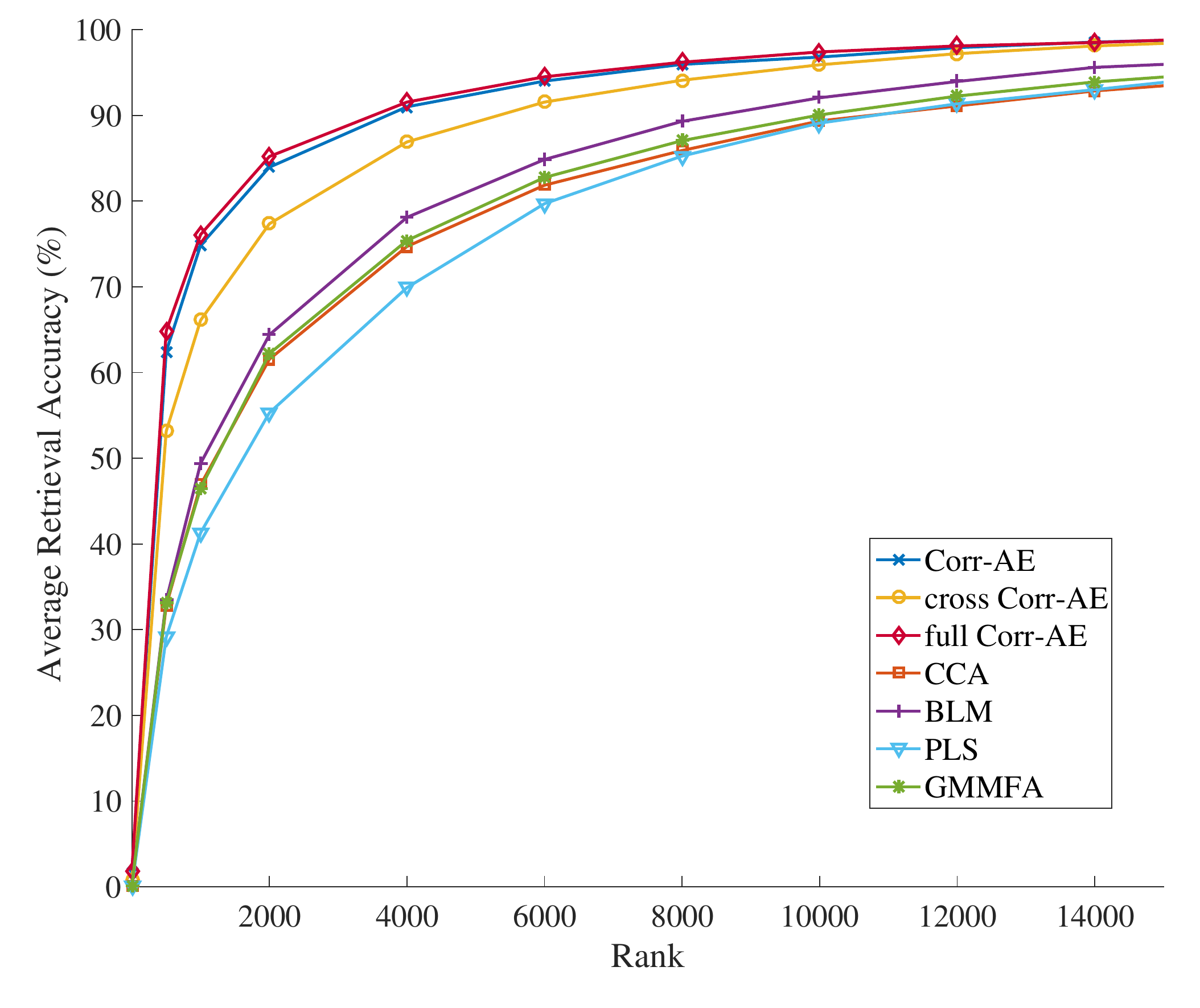}
    \vskip -2mm
    \caption*{Image to text retrieval}
  \end{subfigure}
  \begin{subfigure}{3.5in}
    \caption{Flickr30k}
  \end{subfigure}
  \hfill
  \begin{subfigure}{1.7in}
    \includegraphics[width=1.7in]{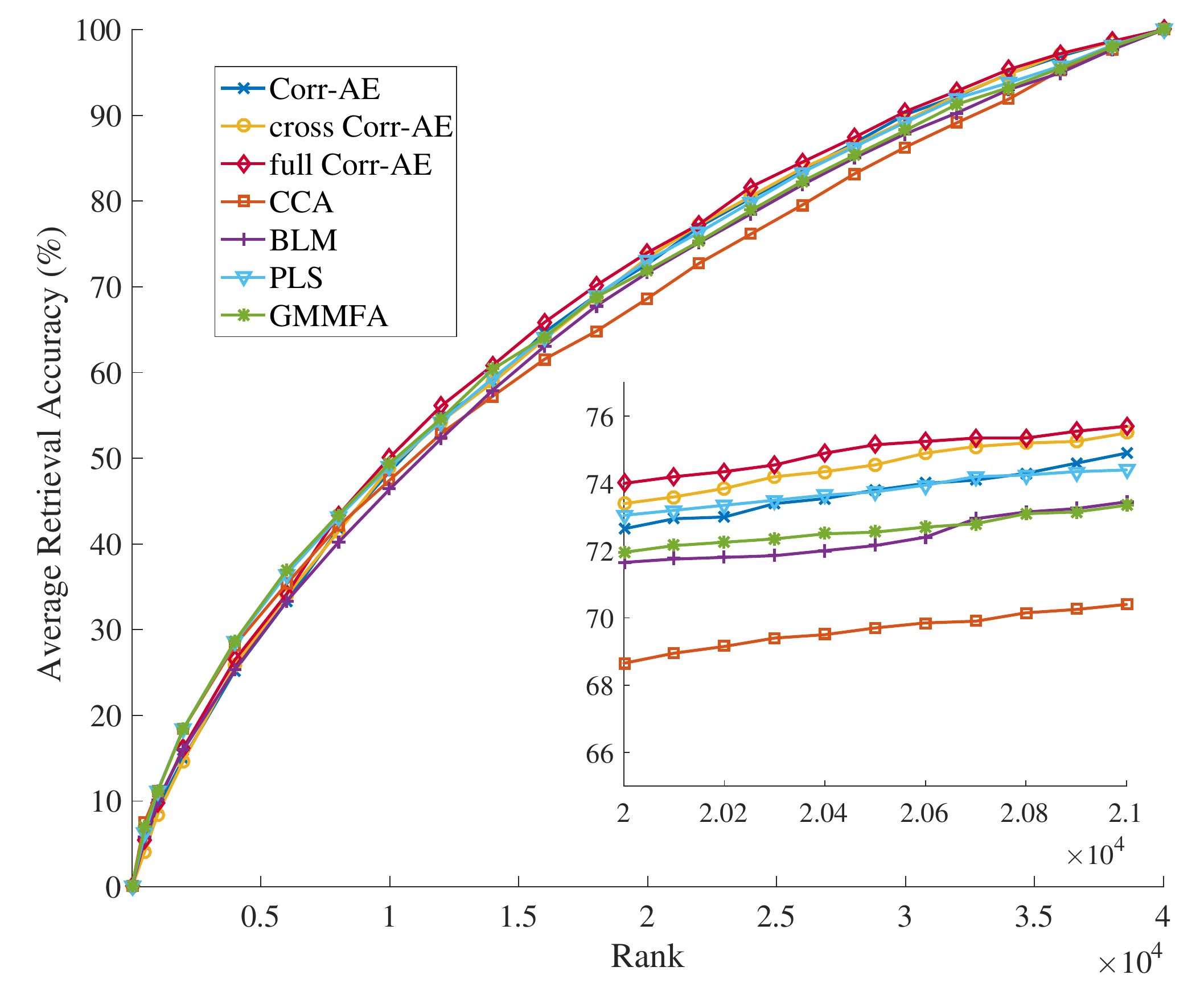}
    \vskip -2mm
    \caption*{Text to image retrieval}
  \end{subfigure}
  \begin{subfigure}{1.7in}
    \includegraphics[width=1.7in]{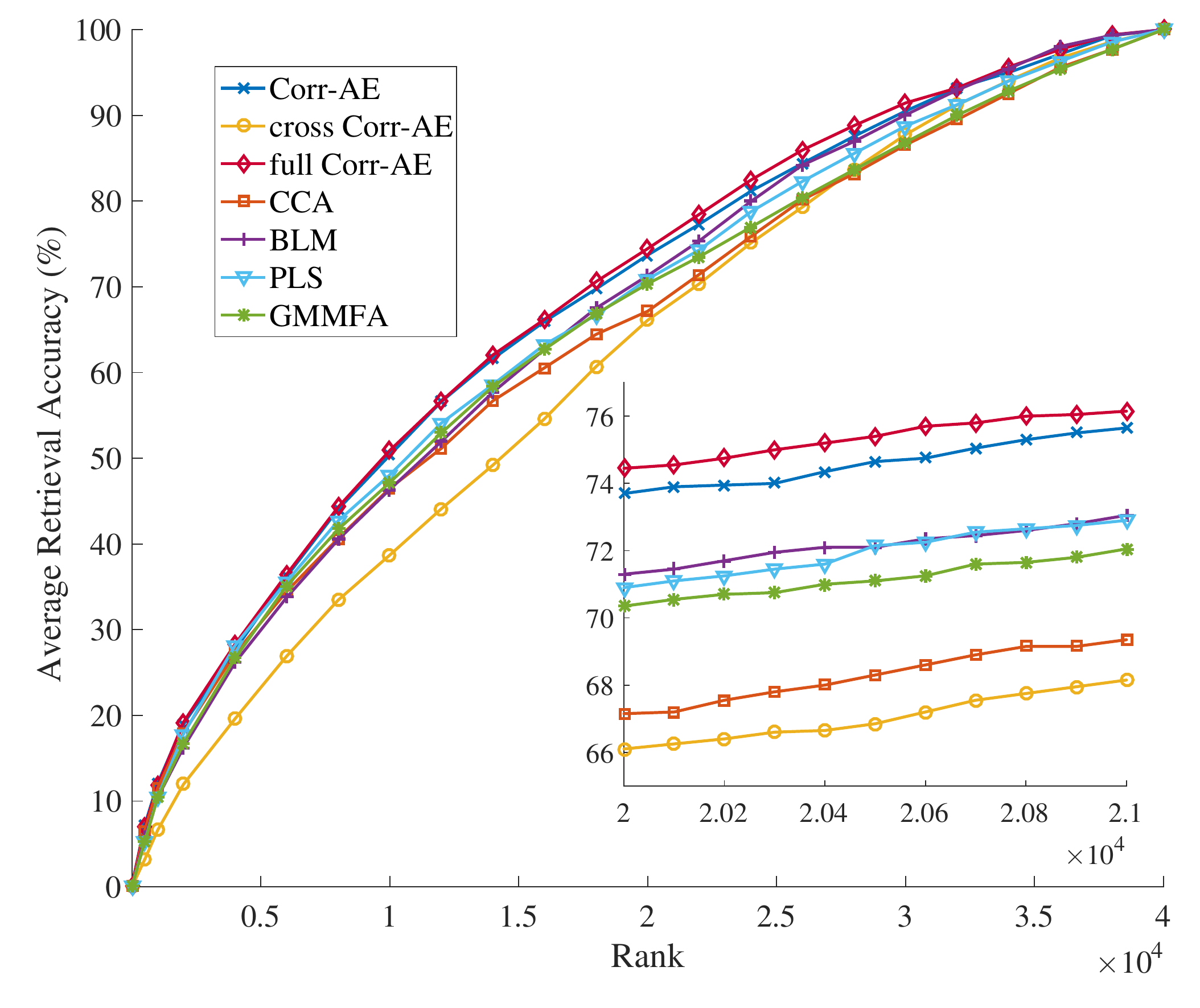}
    \vskip -2mm
    \caption*{Image to text retrieval}
  \end{subfigure}
  \begin{subfigure}{3.5in}
    \caption{Twitter100k}
  \end{subfigure}
  \caption{CMC curves on the (a) Wikipedia, (b) Flicker30k, and (c) Twitter100k datasets. Various Corr-AE and subspace learning methods are compared.}
  \label{dataset_cmc} 
\end{figure}

The benchmarking results on Wikipedia, Flickr30k and Twitter100k are presented in Fig. \ref{dataset_cmc}. Two findings can be drawn.

First, Corr-AE methods surpass the subspace learning methods (CCA, BLM, PLS and GMMFA), especially on the Wikipedia and Flickr30k dataset. The reason is that the subspace learning methods are based on a two-stage framework, which first extracts features for each modality separately and then finds a linear matrix to project two modalities into a shared space. In other words, correlation learning is separated from representation learning. By contrast, Corr-AE methods incorporates representation learning and correlation learning into a single process by defining the loss function as the weighted sum of reconstruction error and correlation loss. Thus Corr-AE methods can achieve superior performance to linear-projection methods. Among the three variants of Corr-AE methods, full Corr-AE achieves the best performance on the whole. In full Corr-AE, two modalities are reconstructed with one modality as input, hence the correlation between images and texts is well embedded in the hidden feature of the autoencoder. Moreover, cross Corr-AE performs poorly in image to text retrieval, which manifests the heterogeneous gap between the two modalities.

Second, generally speaking, the retrieval performance on Flickr30k is the highest among the three datasets while the performance on Twitter100k is the lowest. In other words, the Twitter100k dataset can be viewed as the most challenging one among the three datasets. We speculate that the relatively high performance of the Flickr30k dataset can be attributed to the fact that the texts are written by native English speakers to directly describe the images; the texts and the images are highly correlated. For the Wikipedia dataset, the text-image pairs are collected from Wikipedia's featured articles and the long texts introduce abundant aspects about the iamges, including background, history, episode and so on. Consequently, majority words in the texts are not visual interpretations of the images. The reason why Twitter100k has lowest retrieval accuracy lies in three aspects. First, this dataset covers a diversity of domains. Second, informal expressions often exist in tweets. Third, the texts and images have a relatively loose correlation, and the correlation is even various from user to user. As a result, it is challenging to learn the correlation between the texts and images for Twitter100k.

\subsection{The Impact of Different Amount of Training Data}
\begin{figure}[t]
  \begin{subfigure}{1.7in}
    \includegraphics[width=1.7in]{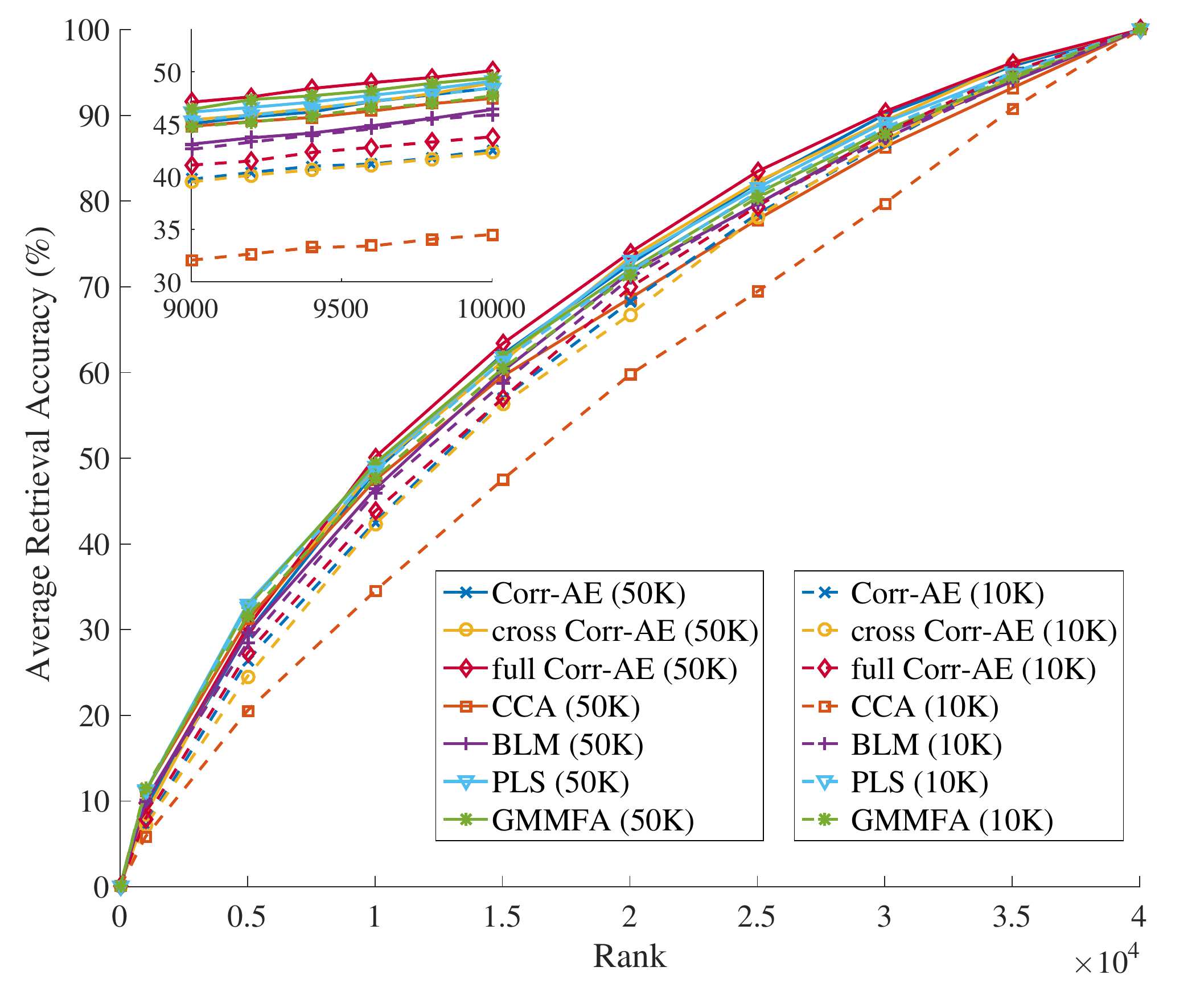}
    \vskip -2mm
    \caption{Text to image retrieval}
  \end{subfigure}
  \begin{subfigure}{1.7in}
    \includegraphics[width=1.7in]{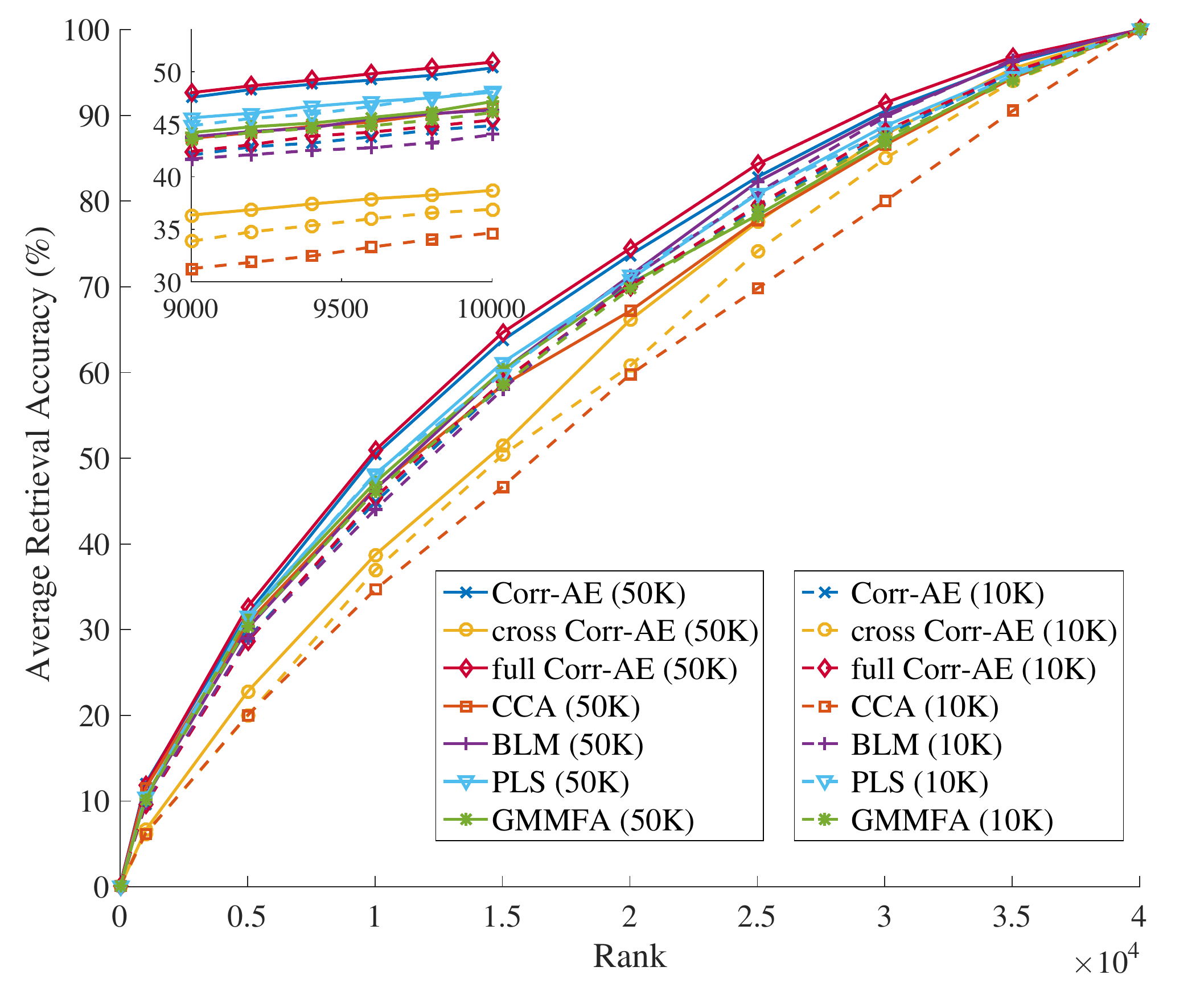}
    \vskip -2mm
    \caption{Image to text retrieval}
  \end{subfigure}
  \caption{CMC curves on the Twitter100k dataset with 10k or 50k image-text pairs as training data.}
  \label{twitter100k_cmc_amount} 
\end{figure}

In this section, we explore whether retrieval performance can be improved with larger amount of training data. To this end, we use 50,000 and 10,000 image-text pairs for training, respectively. The experiment results on Twitter100k dataset with different quantity of training data are given in Fig. \ref{twitter100k_cmc_amount}.

As is shown in Fig. \ref{twitter100k_cmc_amount}, all the dotted lines (10k training data) are below the corresponding solid line (50k training data). For text to image retrieval with full Corr-AE, for instance, average retrieval accuracy improves from 42.55$\%$ to 48.65$\%$ (+6.10$\%$) at rank$=9500$ when the available training data increases from 10k to 50k.

Abundant training data can lead to high ability for generalization and effective correlation learning. The improvement brought by the massive data verifies that the large scale of Twitter100k is a crucial advantage for cross-media retrieval.

\subsection{Comparison of Various Text Features}
\begin{figure}[t]
  \begin{subfigure}{1.7in}
    \includegraphics[width=1.7in]{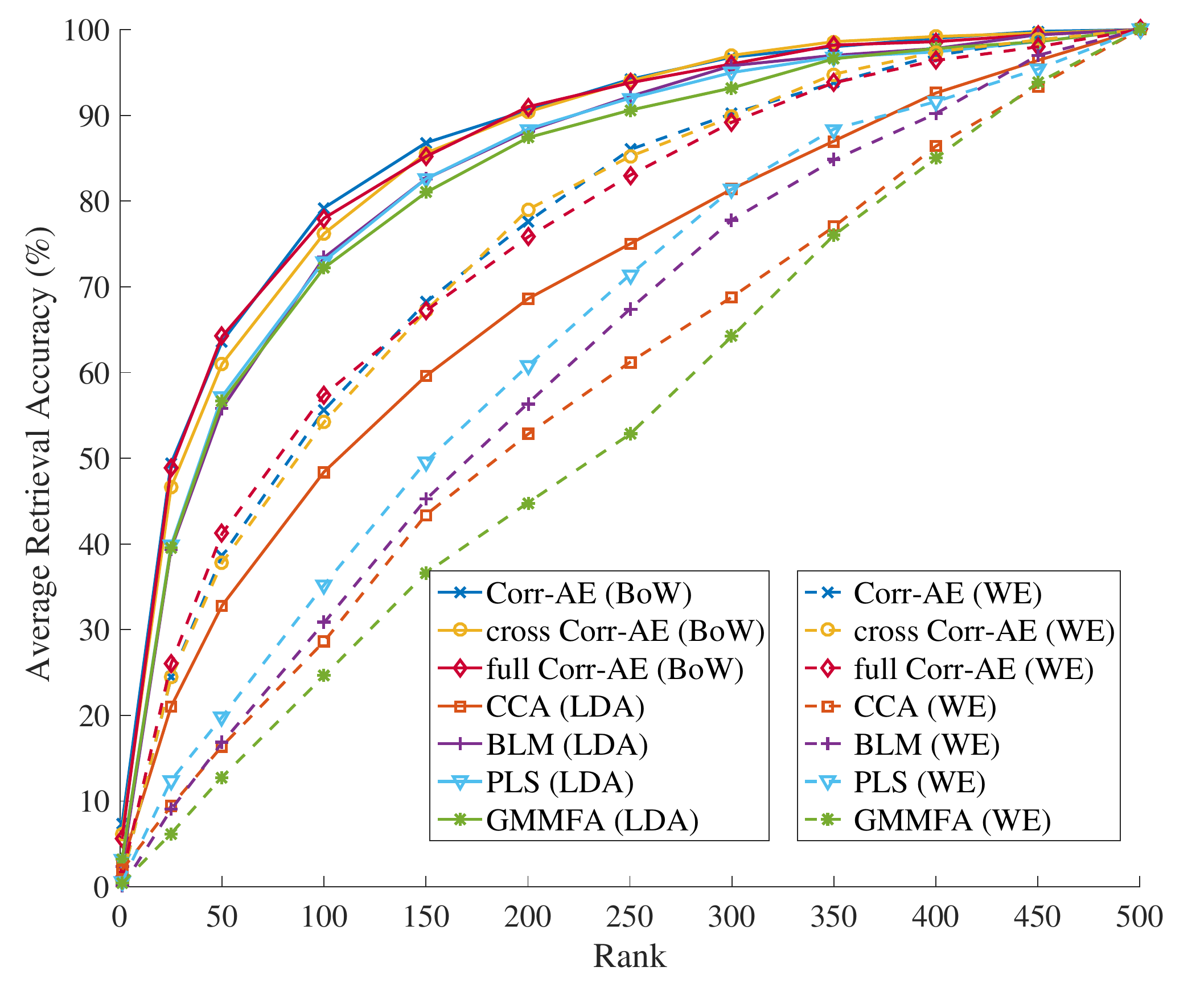}
    \vskip -2mm
    \caption*{Text to image retrieval}
  \end{subfigure}
  \begin{subfigure}{1.7in}
    \includegraphics[width=1.7in]{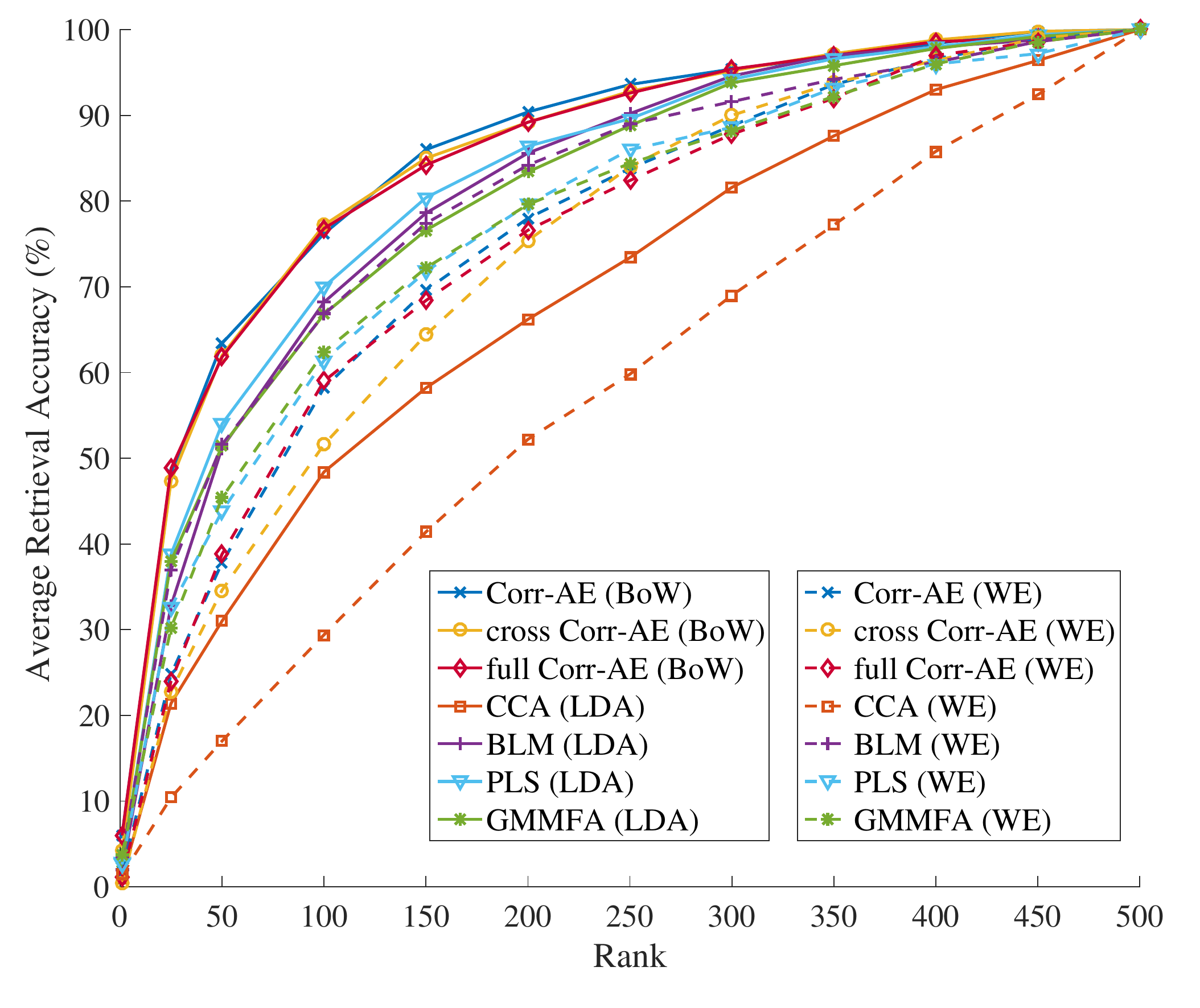}
    \vskip -2mm
    \caption*{Image to text retrieval}
  \end{subfigure}
  \begin{subfigure}{3.5in}
    \caption{Wikipedia}\label{wikipedia_cmc_feature}
  \end{subfigure}
  \hfill
  \begin{subfigure}{1.7in}
    \includegraphics[width=1.7in]{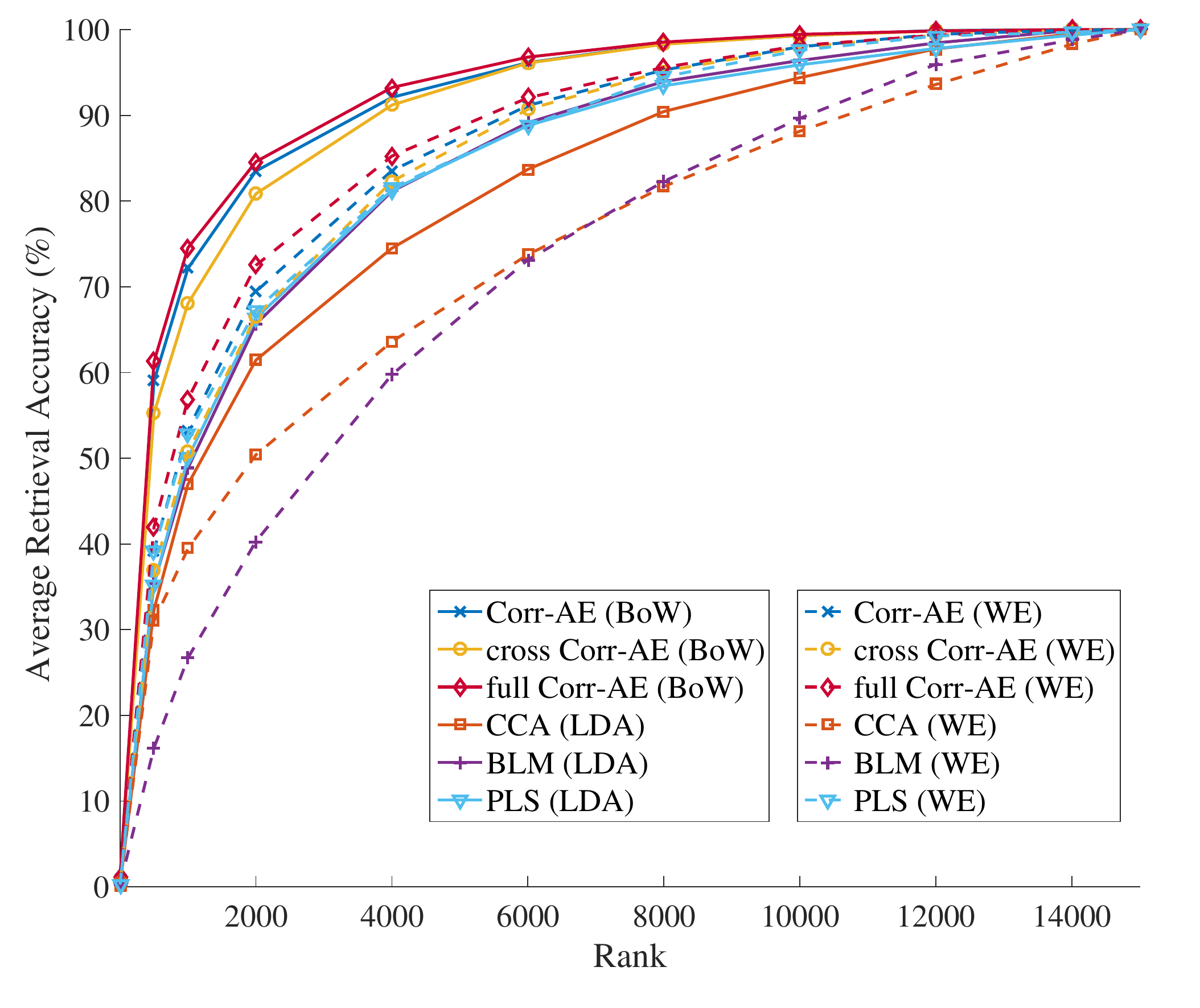}
    \vskip -2mm
    \caption*{Text to image retrieval}
  \end{subfigure}
  \begin{subfigure}{1.7in}
    \includegraphics[width=1.7in]{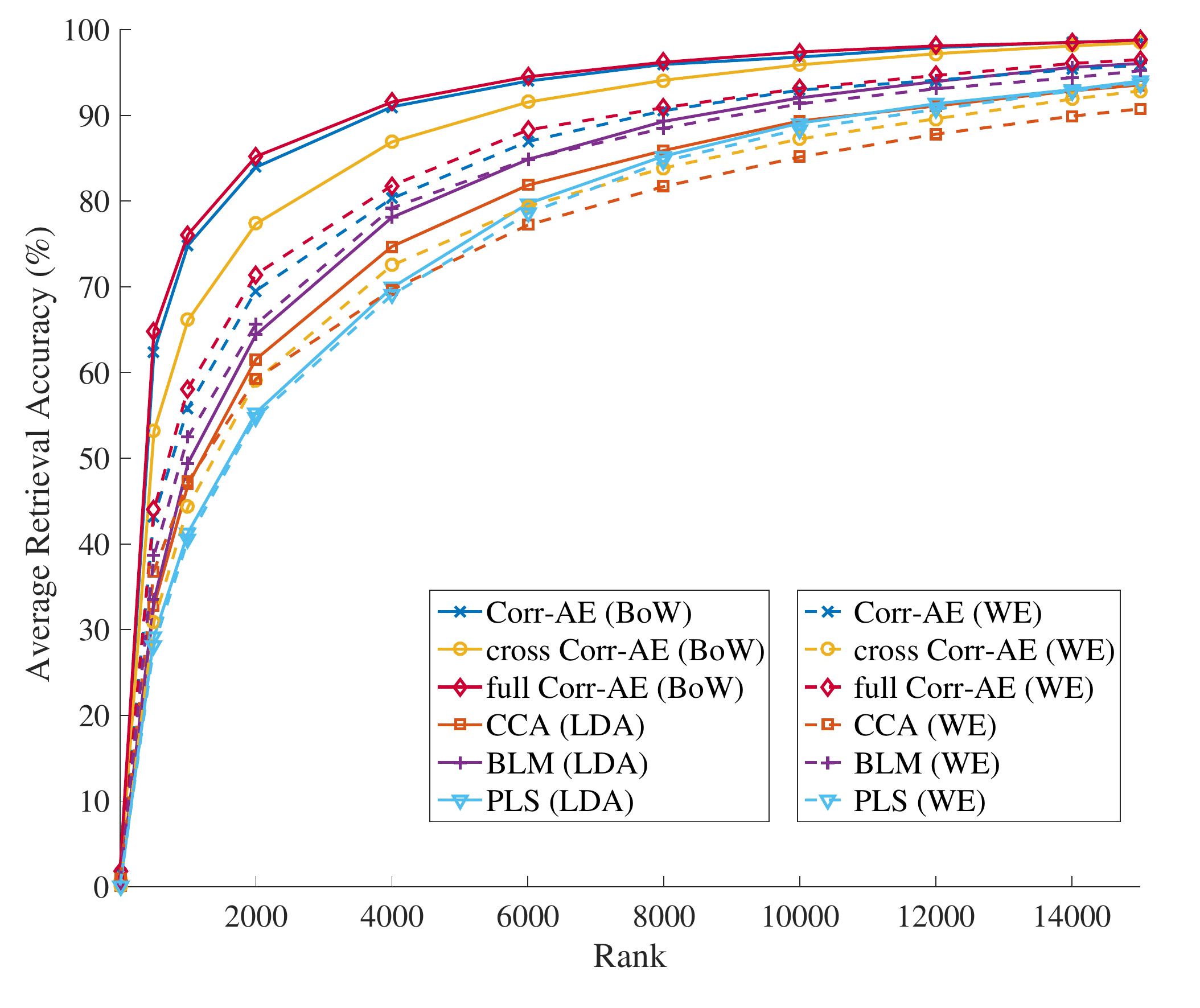}
    \vskip -2mm
    \caption*{Image to text retrieval}
  \end{subfigure}
  \begin{subfigure}{3.5in}
    \caption{Flickr30k}\label{flickr30k_cmc_feature}
  \end{subfigure}
  \hfill
  \begin{subfigure}{1.7in}
    \includegraphics[width=1.7in]{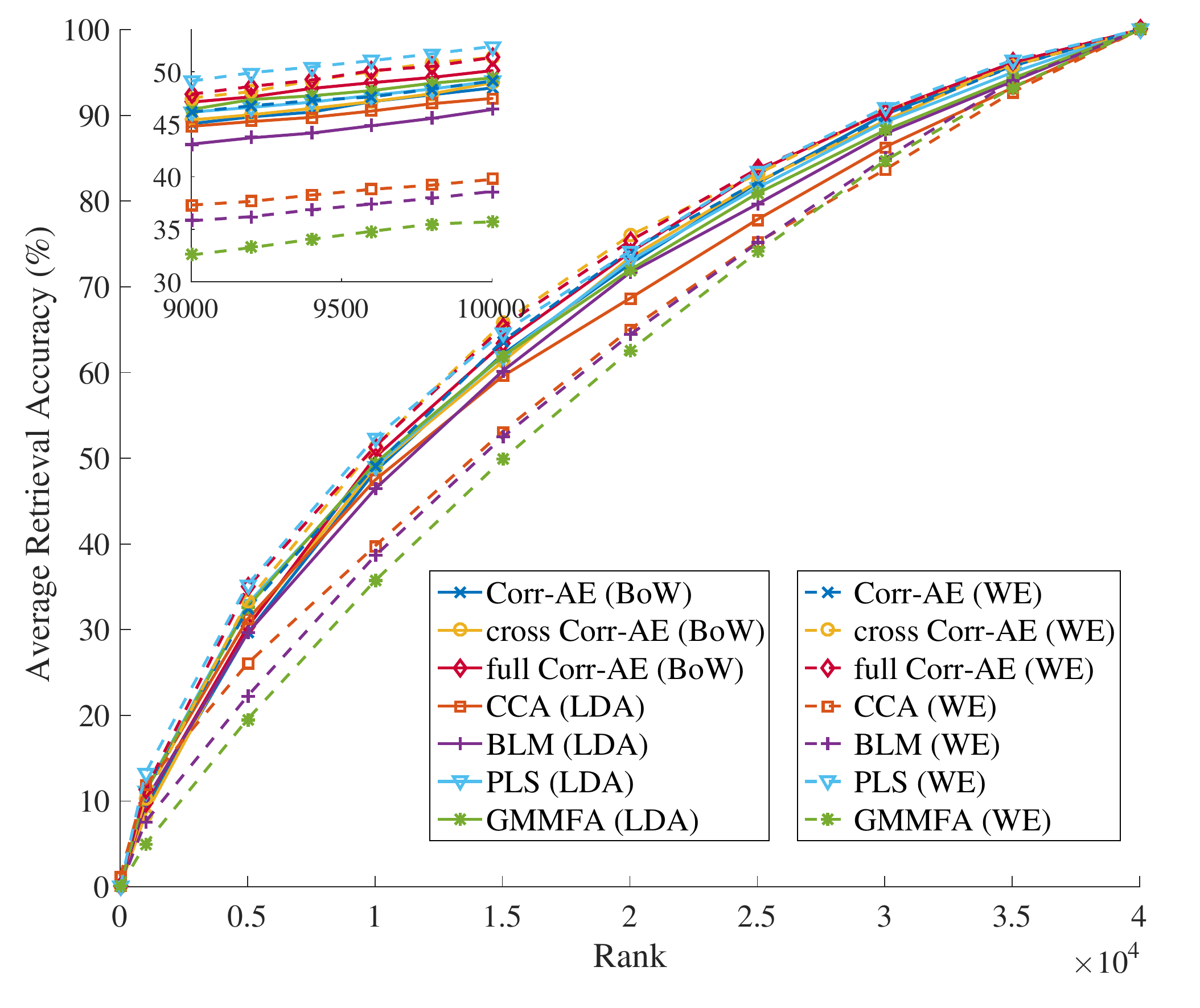}
    \vskip -2mm
    \caption*{Text to image retrieval}
  \end{subfigure}
  \begin{subfigure}{1.7in}
    \includegraphics[width=1.7in]{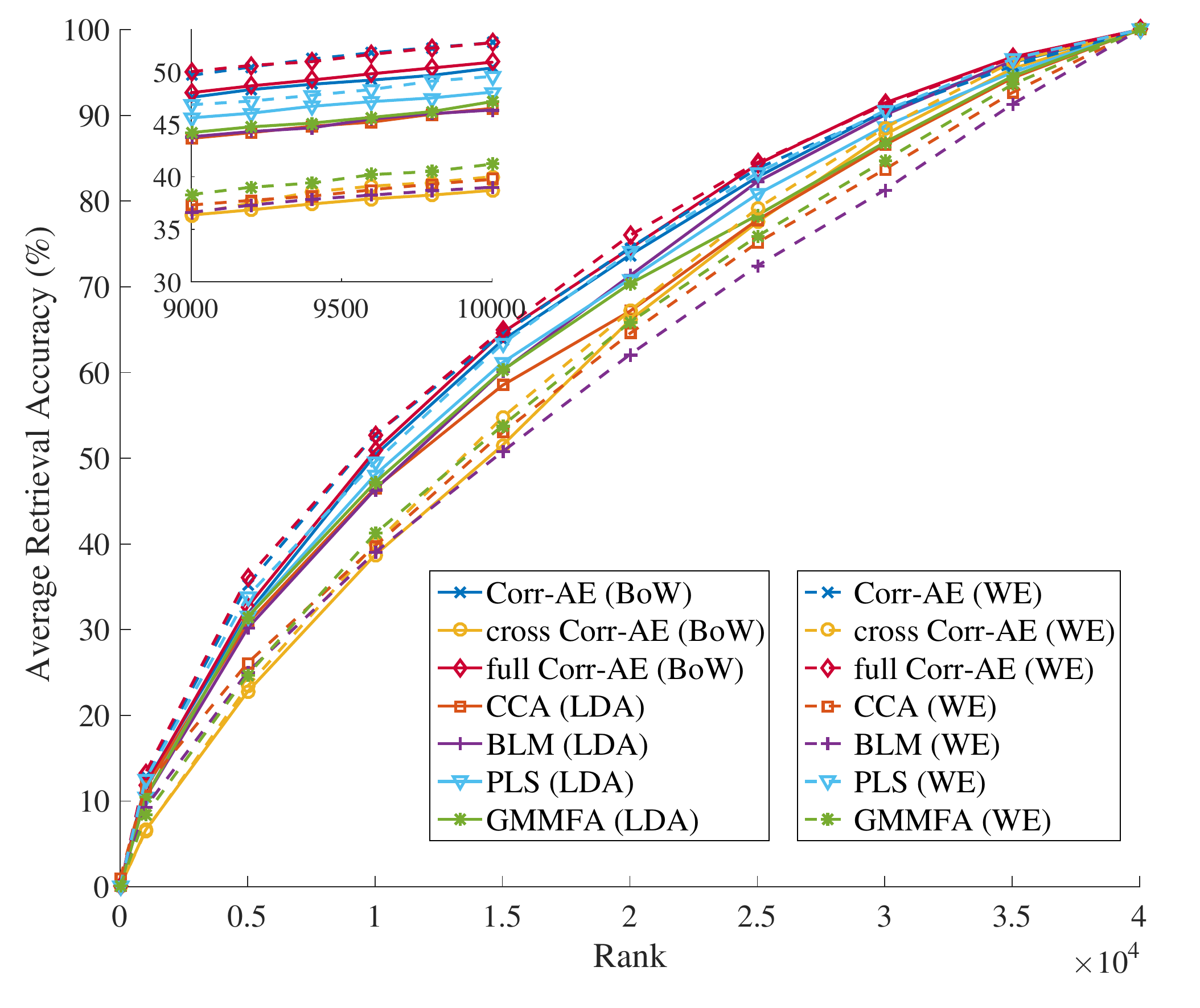}
    \vskip -2mm
    \caption*{Image to text retrieval}
  \end{subfigure}
  \begin{subfigure}{3.5in}
    \caption{Twitter100k}\label{twitter100k_cmc_feature}
  \end{subfigure}
\caption{CMC curves with various text features on the (a) Wikipedia, (b) Flicker30k, and (c) Twitter100k datasets.}
  \label{cmc_feature} 
\end{figure}

To evaluate the influence of different text features, we test the retrieval performance on the Wikipedia, Flickr30k and Twitter100k datasets using various features, \emph{i.e.}, LDA feature, BoW feature and WE-BoW feature. The benchmarking results with different text features are shown in Fig. \ref{cmc_feature}.

The impact of WE-BoW feature varies on datasets and retrieval methods. Take text to image retrieval as example. Similar findings can be obtained in image to text retrieval. First, the results in Fig. \ref{wikipedia_cmc_feature} and Fig. \ref{flickr30k_cmc_feature} demonstrate that the WE-BoW feature is secondary to the LDA feature and BoW feature for cross-media retrieval on Wikipedia and Flickr30k. For example, after exploiting the WE-BoW feature, average retrieval accuracy decreases from 90.60$\%$ to 77.60$\%$ (-13.0$\%$) for Corr-AE at rank$=200$ on Wikipedia. Even more serious deterioration can be seen for BLM at rank$=2000$ on Flickr30k, from 65.65$\%$ to 40.30$\%$ (-25.35$\%$). Second, similar degradations in performance can be observed on Twitter100k when CCA, BLM and GMMFA are adopted, \emph{e.g.}, the accuracy is lower by 7.60$\%$ at rank$=9200$ for CCA. Third, the usage of the WE-BoW feature ameliorates the retrieval accuracy for Corr-AE methods and PLS on Twitter100k. For instance, the accuracy exceeds by +2.65$\%$ and +3.3$\%$ for cross Corr-AE at rank$=9400$ and PLS at rank$=9200$, respectively.

The improvement can be attributed to the semantic information brought by the word vectors, which are trained on large corpus of texts. The vectors of similar words are close in space. In contrast, words are considered independent in the BoW representation. Since texts in Twitter100k contain plenty of informal expressions and abbreviations, semantic information embedded in WE-BoW benefits the retrieval.

However, the text in Wikipedia and Flickr30k gives a detailed description of image and is written in formal language. LDA and BoW are effective to represent text in this kind of corpus. Since WE-BoW is derived from the pre-trained word vectors , LDA and BoW offers an advantage over the WE-BoW feature on Wikipedia and Flickr30k.

On Twitter100k, the opposite effect of word embedding on different retrieval methods results from the merit and drawback of WE-BoW . The merit is the semantic information contained by word vectors. The drawback is the high dimension compared with the 50-dimensional LDA feature. For subspace learning methods, 1024-dimensional WE-BoW makes it difficult to find the shared space for the cross-media data, hence retrieval performance suffers from the WE-BoW feature. On the contrary, since WE-BoW is lower in dimension than the 5000-dimensional BoW feature used for Corr-AE methods, retrieval accuracy is improved with the integration of word embedding.

In addition, the word vectors used in this experiments are pre-trained on text corpus written in well-organized language, specific word vectors trained on Twitter corpus will be more beneficial for the Twitter100k dataset.

\subsection{Performance of the Proposed OCR-based Method}
\begin{figure}[t]
  \begin{subfigure}{1.7in}
    \includegraphics[width=1.7in]{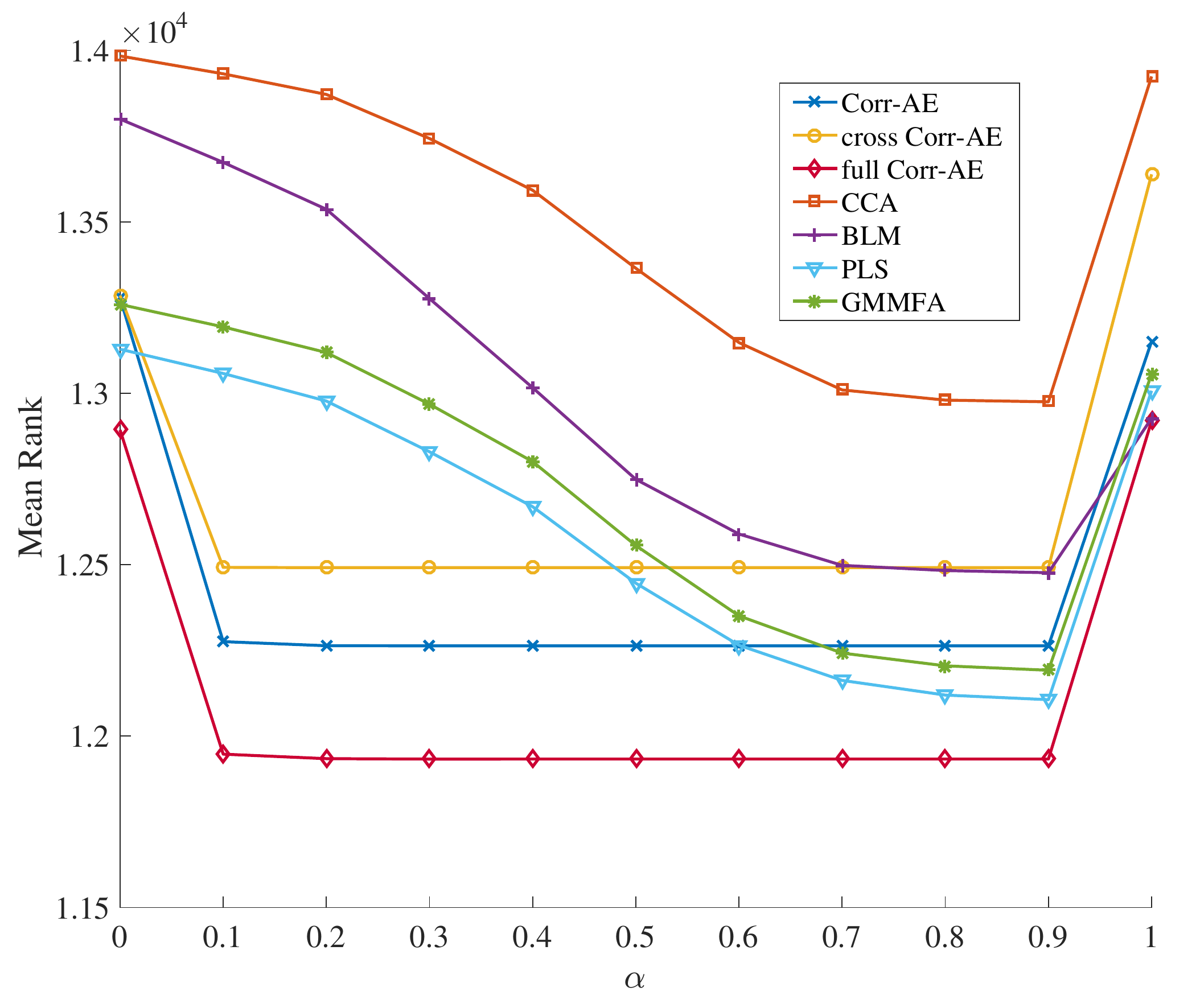}
    \vskip -2mm
    \caption{Text to image retrieval}
  \end{subfigure}
  \begin{subfigure}{1.7in}
    \includegraphics[width=1.7in]{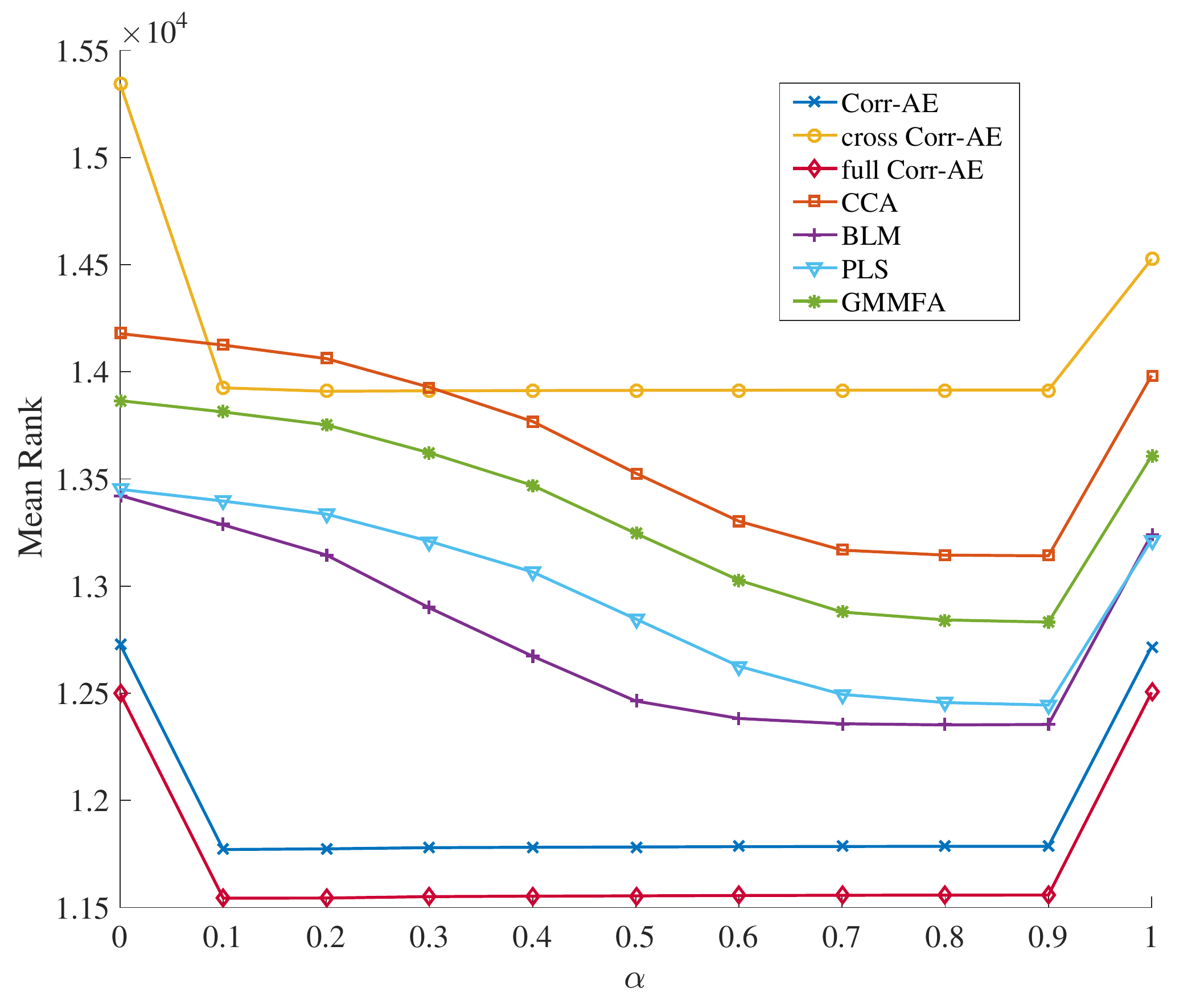}
    \vskip -2mm
    \caption{Image to text retrieval}
  \end{subfigure}
  \caption{Mean rank of the correct matches with different values of $\alpha$ on the Twitter100k dataset.}
  \label{meanR_alpha} 
\end{figure}

This section presents the performance of the proposed OCR-based retrieval method on Twitter100k. Above all, we discuss the impact of the weight parameter $\alpha$ on the retrieval performance. The mean rank of the correct matches with different values of $\alpha$ on Twitter100k are provided in Fig. \ref{meanR_alpha}.

These results reveal that the mean rank of the correct matches declines with $\alpha$ when $\alpha<1$ for all the baselines. It shows that text on image plays a dominant role in retrieval. But the performance deteriorates when $\alpha$ is set to 1, because the information about the color and the shape of the image is lost. We accordingly choose 0.9 as the best value for $\alpha$.

\begin{figure}[t]
  \begin{subfigure}{1.7in}
    \includegraphics[width=1.7in]{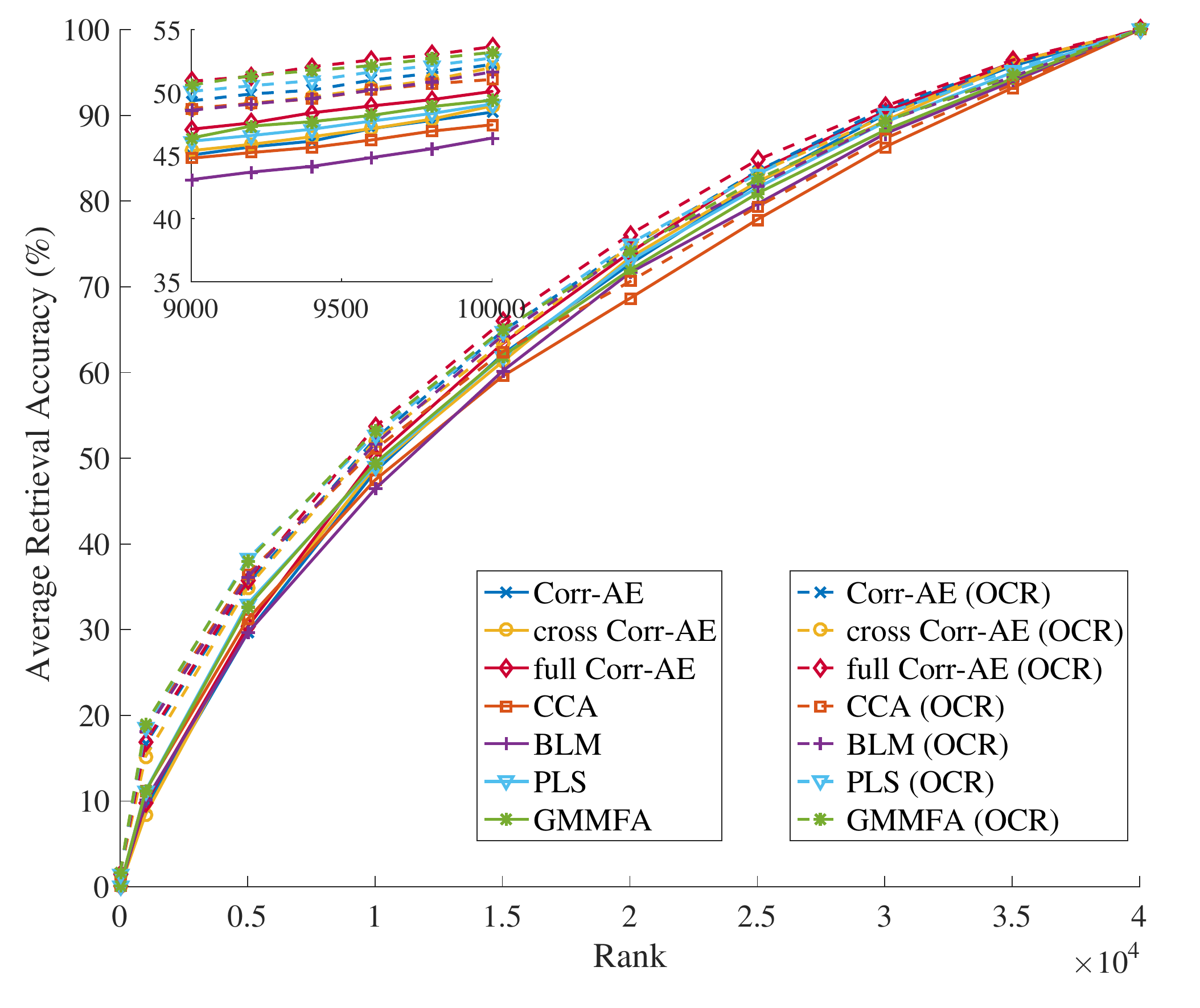}
    \vskip -2mm
    \caption{Text to image retrieval}
  \end{subfigure}
  \begin{subfigure}{1.7in}
    \includegraphics[width=1.7in]{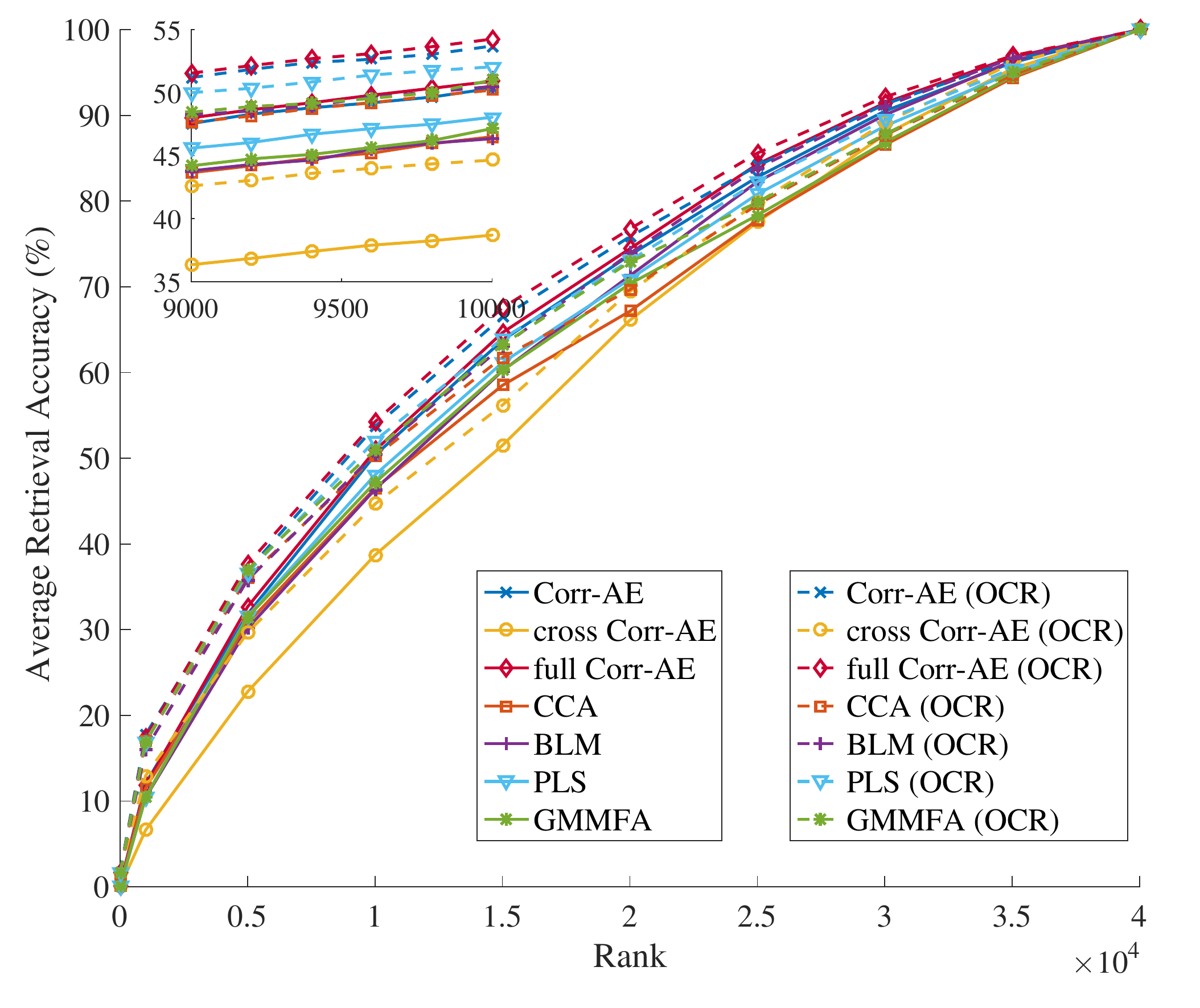}
    \vskip -2mm
    \caption{Image to text retrieval}
  \end{subfigure}
  \caption{CMC on Twitter100k dataset with proposed method and baselines.}
  \label{ocr_cmc} 
\end{figure}

Then we compare the performance of the proposed OCR-based method with baselines. The experiment results are represented in Fig. \ref{ocr_cmc}.

From Fig. \ref{ocr_cmc} we can find out that all the dotted lines (the proposed method) are above the corresponding full line (baselines). In other words, average retrieval accuracy improves by incorporating OCR when the same retrieval method is used, \emph{e.g.}, for BLM in text to image retrieval, average retrieval accuracy ascends from 43.10$\%$ to 48.60$\%$ (+5.50$\%$) at rank$=9000$. This enhancement is ascribed to the facts that the tweets and the texts on the images are highly correlated; the proposed methods can utilize the text information of the images besides the shape and color information.

The proposed OCR-based method may be further developed by utilizing superior similarity metrics and advanced text retrieval methods. Moreover, besides modifying the distance formulation by incorporating Jaccard distance between tweet and OCR-text, OCR-text can be integrated in multiple ways, such as concatenating the OCR-text with other image features and so forth.

\subsection{Retrieved Examples of the Twitter100k Dataset}
In this section, we study the retrieval results of the Twitter100k dataset. We adopt full Corr-AE method and BoW feature. OCR-texts and 50,000 training data are employed. For each query, we represent top five retrieved results in the rank list and provide the ground truth match together with its rank for reference.

\begin{figure*}[t]
  \begin{subfigure}{7.2in}
    \includegraphics[width=7.2in]{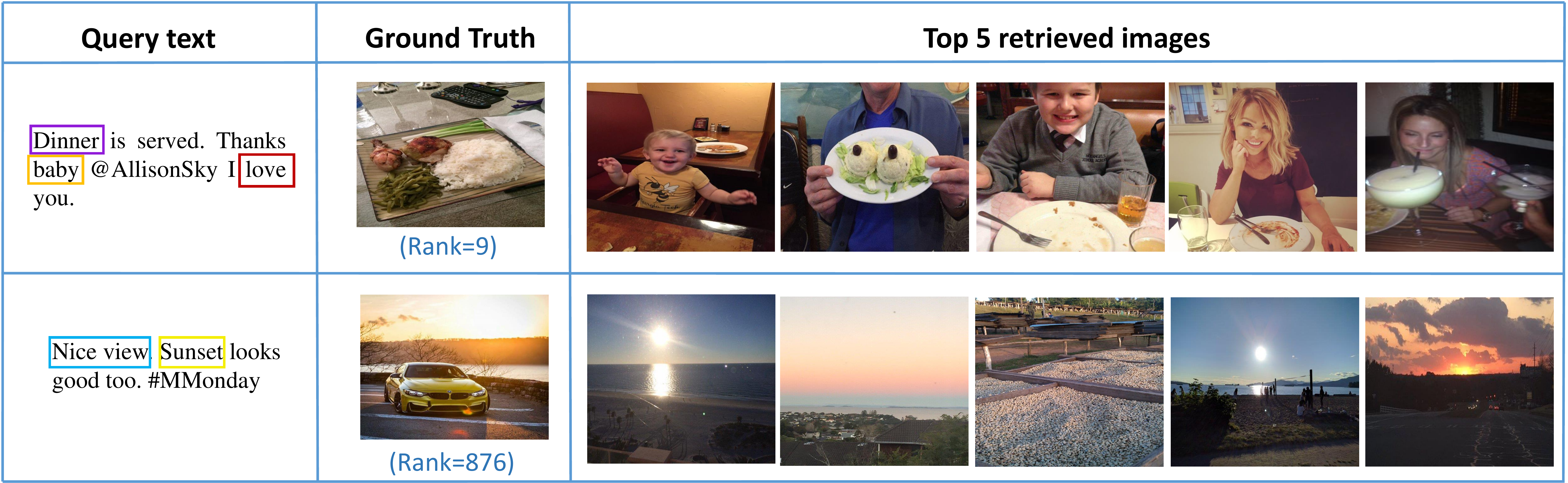}
    \caption{Text to image retrieval}
  \end{subfigure}
  \hfill
  \vskip 2mm
  \begin{subfigure}{7.2in}
    \includegraphics[width=7.2in]{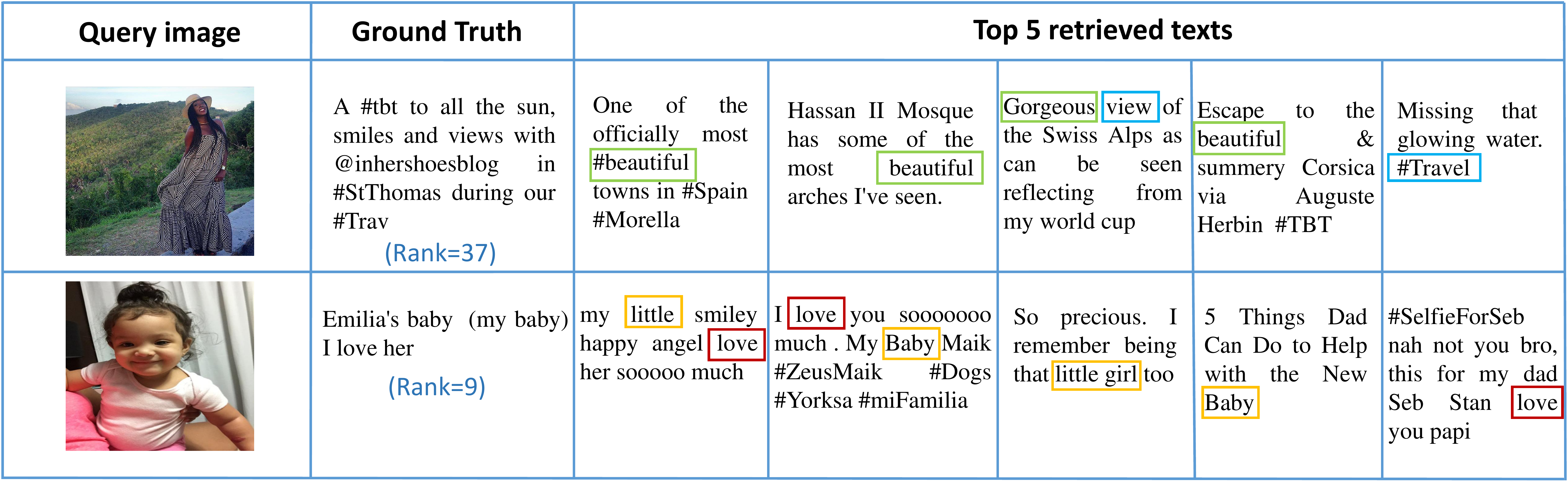}
    \caption{image to text retrieval}
  \end{subfigure}
  \caption{Top retrieved examples of the full Corr-AE method on the Twitter100k dataset. The key words which are correlated to the images from the perspective of content and opinion are marked with rectangles.}
  \label{top_retrieved_example} 
\end{figure*}

As is demonstrated in Fig. \ref{top_retrieved_example}, the ground truth match and top five retrieved results are correlated to the query from two perspectives. One is the content aspect. Take the first text to image retrieval for example. \emph{Dinner} and \emph{baby} are two key words in the query text. The food and drinks in images are the visual interpretations of \emph{dinner}. The people in images are corresponding to \emph{baby} because \emph{baby} can be used as hypocorism to call the person loved by someone besides infant.

Another perspective is opinion or sentiment. For instance, in the first image to text retrieval, the query image consists of a woman in dress, sky, clouds, grass, trees and mountains. It is a scenic image taken during the travel, which can be confirmed by the ground truth text. Opinion words such as \emph{beautiful} and \emph{gorgeous} appear in the top four retrieved texts, which manifest the sentiment of the Internet users upon the scenery and travel.

In Twitter100k dataset, image is more than what it shows. It involves opinion and sentiment of the Internet users in addition to the content. Cite the second image to text retrieval as an example, with the query image of a girl, in addition to \emph{little} and \emph{baby}, \emph{love} is a high frequency word in the retrieved texts. This implies that opinion aspect has an influence on multi-media retrieval although the sentiment of an image is hidden in a high semantic level.

\section{Conclusion}
In this paper, we introduce a large-scale cross-media dataset called Twitter100k, which provides a more realistic benchmark towards weakly supervised text-image retrieval. A number of learning methods and text features are evaluated in our experiments. Considering the characteristic of the new dataset, we propose to improve the retrieval performance based on OCR-texts of images.

There is still a long way to go before achieving a satisfying retrieval performance on the new dataset. In future work, we plan to design a better evaluation protocol for this dataset and consider the correlation between the images and texts from high-level semantic perspectives, such as opinion and sentiment views. We expect this paper and the new dataset will motivate more insightful works.

\section*{Acknowledgment}
This research is supported by the Key Program of National Natural Science Foundation of China (Grant No. U1536201 and No. U1536207).

\ifCLASSOPTIONcaptionsoff
  \newpage
\fi

\bibliographystyle{IEEEtran}
\bibliography{Twitter100k}

\begin{IEEEbiography}[{\includegraphics[width=1in,height=1.25in,clip,keepaspectratio]{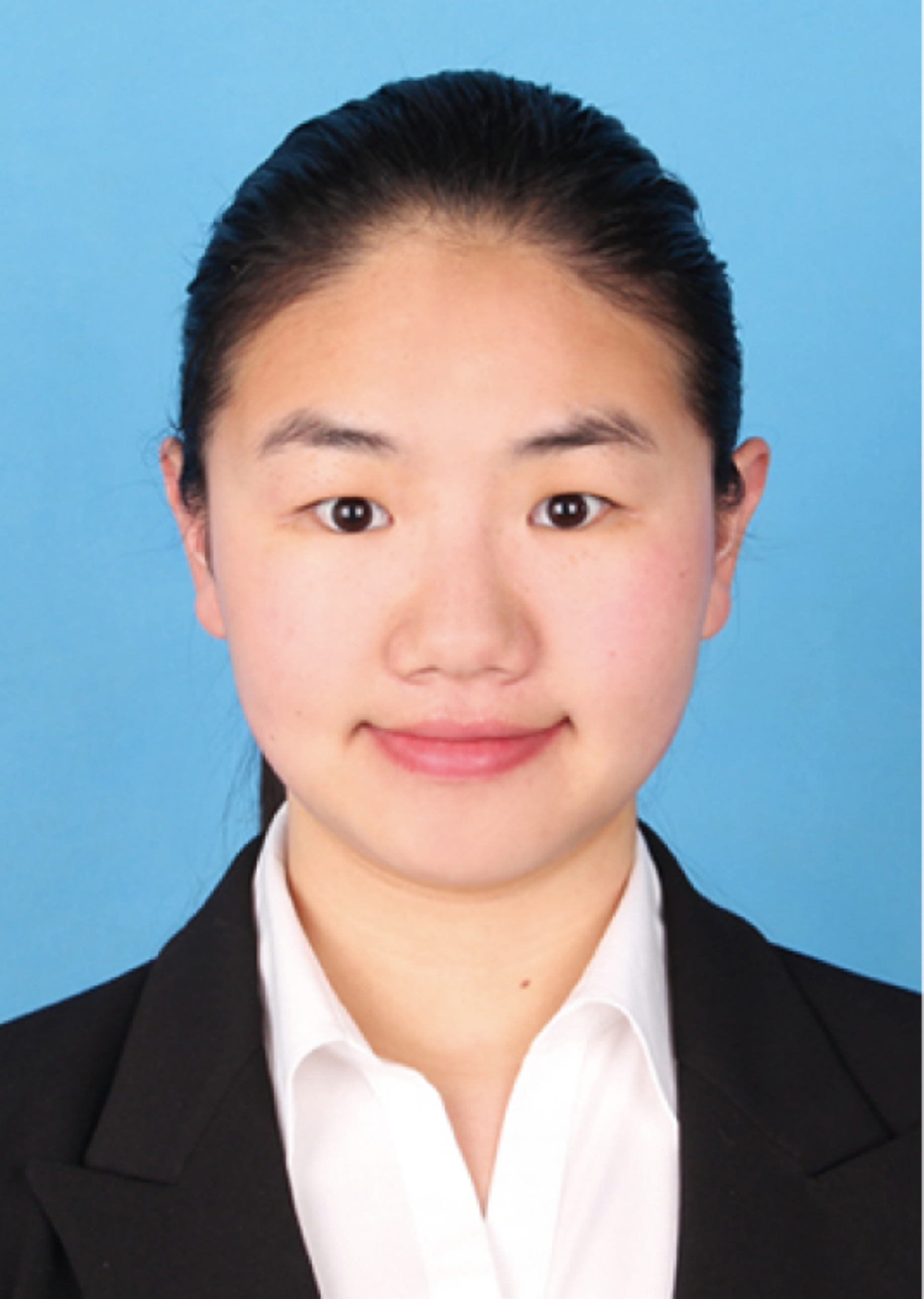}}]{Yuting Hu} received the B.E. degree in electronic engineering in 2016 from Tsinghua University, Beijing, China, where she is currently working toward Ph.D. degree. Her current research interests include multimedia information retrieval and natural language processing.
\end{IEEEbiography}

\begin{IEEEbiography}[{\includegraphics[width=1in,height=1.25in,clip,keepaspectratio]{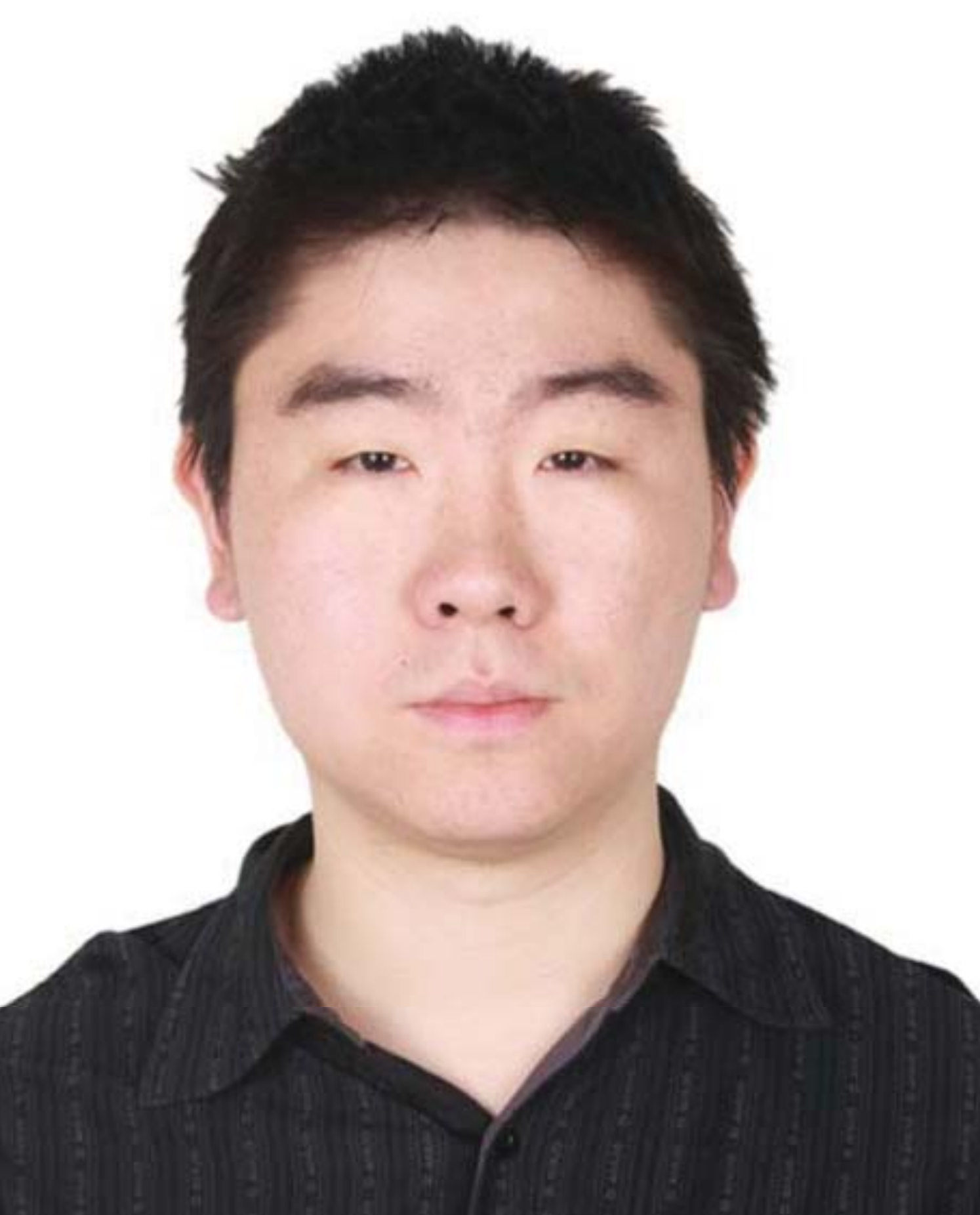}}]{Liang Zheng} received the B.E. degree in life science from Tsinghua University, Beijing, China, in 2010, and the Ph.D. degree in electronic engineering from Tsinghua University in 2015. From August 2015 to June 2016, he was a Postdoctoral Fellow in University of Texas at San Antonio. He is currently a Postdoctoral Fellow in University of Technology Sydney. His research interests include multimedia information retrieval and computer vision.
\end{IEEEbiography}

\begin{IEEEbiography}[{\includegraphics[width=1in,height=1.25in,clip,keepaspectratio]{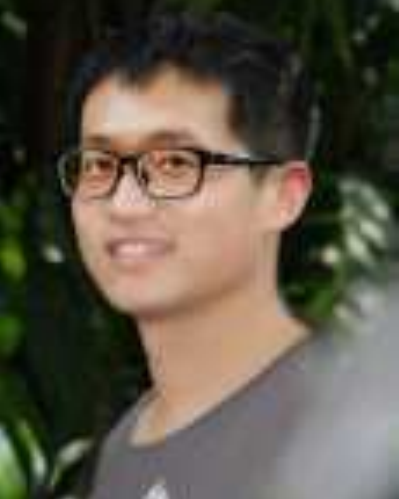}}]{Yi Yang} received the Ph.D. degree in computer
science from Zhejiang University, Hangzhou, China, in 2010. He is currently an associate professor with University of Technology Sydney, Australia.
He was a Post-Doctoral Research with the School of Computer Science, Carnegie Mellon University, Pittsburgh, PA, USA. His current research interest includes machine learning and its applications to multimedia content analysis and computer vision.
\end{IEEEbiography}

\begin{IEEEbiography}[{\includegraphics[width=1in,height=1.25in,clip,keepaspectratio]{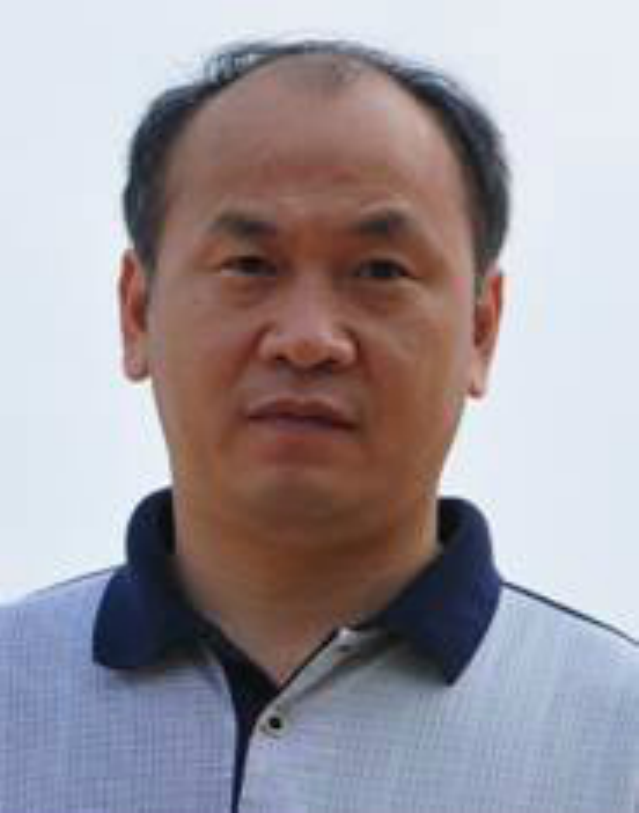}}]{Yongfeng Huang} received the Ph.D degree from Huazhong University of Science and Technology, Wuhan, China, in 2000. From 2000 to 2002, he was a Postdoctoral Fellow with the Department of Electronic Engineering, Tsinghua University, Beijing, China, where he is currently a professor. His research interests include information retrieval, data mining, multimedia network security and next-generation Internet.
\end{IEEEbiography}

\vfill

\end{document}